%% file: acl_latex.tex
\documentclass[11pt]{article}

\usepackage[final]{acl}

\usepackage{times}
\usepackage{latexsym}

\usepackage[T1]{fontenc}

\usepackage[utf8]{inputenc}

\usepackage{microtype}

\usepackage{inconsolata}

\usepackage{multirow}
\usepackage{amsmath}
\usepackage{amssymb}
\usepackage{booktabs}
\usepackage{rotating} 
\usepackage{xcolor}
\usepackage{url}
\usepackage{listings}
\usepackage{array}
\usepackage{graphicx}
\usepackage{xspace}
\usepackage{caption}
\usepackage{subcaption}
\usepackage{wrapfig}
\usepackage[breakable]{tcolorbox}
\usepackage{enumitem}
\usepackage{colortbl}
\usepackage{hyperref}
\usepackage{needspace}
\usepackage{placeins}
\usepackage[dvipsnames,svgnames,x11names,table]{xcolor}

\newcommand{\std}[1]{{\scriptsize$\pm$#1}}
\newcommand{\tableindent}{\hspace{1em}}
\newcommand{\hlcellr}{\cellcolor{BrickRed!10}}
\newcommand{\hlcellg}{\cellcolor{ForestGreen!10}}
\newcommand{\hlcellrr}{\cellcolor{BrickRed!20}}
\newcommand{\hlcellrrr}{\cellcolor{BrickRed!35}}
\newcommand{\hlcellrrrr}{\cellcolor{BrickRed!50}}

\definecolor{lightergray}{HTML}{E9E9E9}
\definecolor{lightblue}{RGB}{190, 210, 230}
\definecolor{lavender}{RGB}{200, 190, 230}

\title{Make an Offer They Can't Refuse: Grounding Bayesian Persuasion in Real-World Dialogues without Pre-Commitment}

\author{
  \textbf{Buwei He\textsuperscript{1,2}\footnotemark[1]},
  \textbf{Yang Liu\textsuperscript{2}\footnotemark[1]},
  \textbf{Zhaowei Zhang\textsuperscript{2,3}},
  \textbf{Zixia Jia\textsuperscript{2}},
  \textbf{Yang Yu\textsuperscript{4}},
  \textbf{Huijia Wu\textsuperscript{1}},
\\
  \textbf{Zhaofeng He\textsuperscript{1}\footnotemark[2]},
  \textbf{Zilong Zheng\textsuperscript{2}\footnotemark[2]},
  \textbf{Yipeng Kang\textsuperscript{2}\footnotemark[2]}
\\
  \textsuperscript{1}Beijing University of Posts and Telecommunications \\
  \textsuperscript{2}Beijing Institute for General Artificial Intelligence \\
  \textsuperscript{3}Peking University 
  \textsuperscript{4}China University of Petroleum (Beijing)
\\
  \small{\{hebuwei, zhaofenghe\}@bupt.edu.cn},
  \small{\{liuyang, zlzheng, kangyipeng\}@bigai.ai}
}

\begin{document}
\maketitle

\renewcommand{\thefootnote}{\fnsymbol{footnote}}
\footnotetext[1]{Equal contribution.}
\footnotetext[2]{Corresponding author.}

\begin{abstract}
Large language models (LLMs) still struggle with strategic persuasion, largely because existing approaches either neglect information asymmetry or rely on unrealistic pre-commitment assumptions. 
We introduce a type-induced commitment-communication mechanism that grounds Bayesian Persuasion (BP) in natural language dialogue without pre-commitment: the persuader narrates their potential types (e.g., honest vs. dishonest) to dynamically construct an information schema, enabling the persuadee to perform Bayesian belief updates within the conversation itself. We implement two variants: Semi-Formal-Natural-Language (SFNL) and Fully-Natural-Language (FNL), evaluating them against strong baselines across multiple LLMs and human judges. 
BP strategies consistently outperform baselines: SFNL excels in logical credibility, while FNL shows superior robustness and emotional resonance. We verify that gains stem from genuine Bayesian reasoning rather than superficial formatting, and we further show that supervised fine-tuning enables small models to match the persuasive performance of much larger ones.

\end{abstract}

\section{Introduction}

\input{figures/teaser}

Persuasion is a fundamental form of human social interaction, enabling individuals to influence others' beliefs and decisions through communication~\citep{brembeck_persuasion_1976}. While large language models (LLMs) already exhibit strong abilities in language generation and understanding, they remain limited in strategic persuasion tasks: LLMs fail to design messages that rationally shift a persuadee's beliefs. 

Bayesian Persuasion (BP), a game-theoretic framework for information design, offers a tractable solution in constrained mathematical settings by modeling how a persuader can disclose information to maximize desired actions~\citep{kamenica_bayesian_2011}. However, applying BP in open-ended natural language dialogue raises a key challenge: the mathematical constructs of BP---such as priors, world states, and posterior updates---must be \emph{verbalized} into coherent and persuasive arguments. 
Moreover, a pivotal step in BP is the persuader's commitment to a \emph{signaling schema}, which we argue is crucial for natural language realization. Existing approach relies on \textbf{pre-commitment} by statically encoding the schema in the persuadee's prompt, bypassing the need for its communication~\citep{li2026verbalized}. 

To overcome the limitation, we introduce a type-induced commitment-communication mechanism: rather than assuming pre-commitment, the persuader explicitly narrates their potential types (e.g., honest or dishonest) during the communication. This verbal articulation of the information schema enables the persuadee to perform Bayesian posterior updates. Thus, \textbf{we recast the schema pre-commitment as type disclosure}, bridging the gap between formal BP theory and authentic natural language implementation.


As illustrated in Figure~\ref{fig:teaser}, we implement the commitment-communication mechanism with two variants: Semi-Formal-Natural-Language (SFNL) and Fully-Natural-Language (FNL).
We comprehensively evaluate diverse persuadees, including LLM instances with varying prompts and fine-tuning and human participants, across tailored persuasion scenarios and everyday contexts. 

In summary, we propose a framework for grounding BP in natural dialogues, leveraging type-induced commitment to overcome pre-commitment constraint. By systematically evaluating, we provide insights into how LLMs can harness information asymmetry for strategic persuasion. 

\section{Related work}
\subsection{Persuasion with Large Language Models.}
LLMs have shown powerful semantic processing capabilities~\citep{NEURIPS2025_d0251bbb, 10.65109/ZJJG5330, kang2020incorporating} and broad value alignment potential~\citep{zhang2026bach, ziheng-etal-2026-simple, kang-etal-2025-values, shang2026chaos} in social scenarios~\citep{ICLR2026_cbcbf8fe, ziheng-etal-2026-moral, chen-etal-2026-jurisbench} and multi-agent settings~\citep{he2026credibility, mao2025ibgp}, leading to a surge in research on their persuasive capabilities~\citep{zhang2026pila, shi_persuasion_2025, ICLR2026_5f7a70be}.
Additionally, several studies are focused on building specialized  persuasion datasets~\cite{zhang2026naiad, wang-etal-2019-persuasion, jin-etal-2024-persuading,hayati-etal-2020-inspired}. 
CToMPersu~\cite{11464608} proposes a 
framework where persuasion strategies and mental states remain undisclosed, which is well-suited for BP reasoning due to its explicit information asymmetry design.


\subsection{Bayesian Persuasion and Its Verbalization.}
As an information design framework, BP models how an informed sender can strategically disclose information to influence a receiver's beliefs and actions~\citep{kamenica_bayesian_2011}. 
This foundation has been expanded through the broader lens of information design~\cite{bergemann_information_2016,bergemann_information_2019, dughmi_algorithmic_2017, castiglioni_online_2020, bernasconi_optimal_2023, bacchiocchi_online_2024,shaki_bayesian_2025,wojtowicz_when_2024}.

Beyond theoretical extensions, BP has been applied to practical AI challenges. \citet{bai_efficient_2024} employ BP for model-agnostic alignment, demonstrating BP's utility in improving model performance without direct training. 
\citet{zhang2025roadmap} introduce BP into AI governance, proposing that AI can be made to act according to human intentions by designing information.

Recent work has begun to operationalize BP in natural language. \citet{li2026verbalized} propose Verbalized Bayesian Persuasion (VBP), which incorporates the signaling schema directly into the persuadee's prompt and adjusts persuader generation via keyword manipulation. Although VBP works well in constrained settings, it requires the persuadee to know the schema prior to the dialogue. This pre-commitment assumption limits its applicability to open-ended, realistic conversations.

In contrast, our method enables \textbf{fully self-derived schema communication}: the persuader explicitly articulates the information structure within the natural language discourse itself. This commitment-communication mechanism allows our approach to be model-agnostic, and generalizable across diverse scenarios using a unified prompting strategy. 

\section{Type-Induced BP in Natural Language}

In this section, we consider one-turn, two-agent persuasion scenarios for our approach. Diverging from the classical BP model, our framework relaxes the assumption of prior commitment to common knowledge. Instead, we exploit the richness of natural language, allowing the persuader to shape this common knowledge through dynamic interaction. By integrating BP logic into persuasion, our approach achieves higher success rate.


\subsection{General Framework}\label{subsec:General_Framework}
The general framework comprises the following components:
\begin{itemize}[leftmargin=*,nosep]
    \item \textbf{Players:} A Persuader (S) and a Persuadee (R).
    \item \textbf{State of the World:} A finite set of possible states $\Omega = \{\omega_1, \dots, \omega_K\},\quad K \ge 2$. The true state $\omega \in \Omega$ is observable to S but not to R.
    \item \textbf{Prior Beliefs:} R and S hold prior beliefs $ \mu_R, \mu_S \in \Delta(\Omega) $ about the world state, respectively. These priors are not assumed identical or constitute common knowledge.
    \item \textbf{Actions and Utilities:} S sends a natural language message $m$. R observes $m$ and chooses an action $a \in \{\text{Accept, Reject}\}$. R aims to maximize its expected utility $u(a, \omega)$, while S aims to maximize the probability of acceptance, $P(a=\text{Accept}\mid m)$.
\end{itemize}

\input{figures/main_fig}
\subsection{The Type-Induced Signal}\label{sec:typetype}
In contrast to the classic BP assumption of a pre-committed common-knowledge information schema $\pi_S(m\mid \omega): \Omega\to\Delta(M)$, where $M$ is a limited set of signal categories, our approach enables dynamic schema conveyance within natural language \emph{type disclosure}.

\paragraph{Composite Signal Structure:} Message $m$ integrates four functional components:
\begin{itemize}[leftmargin=*,nosep]
    \item $m_{\text{basic}}$: Basic background information about $\Omega$, to align the players' understanding.
    \item $m_{\text{type}}$: A narrative regarding the type of S (defined below), employed to construct the underlying information schema.
    \item $m_{\text{des}}$: A description of the observed state $\omega$.
    \item $m_{\text{inf}}$: An explicit inference, guiding R to calculate its expected payoff and conclude that accepting is the optimal action.
\end{itemize}
\paragraph{Type-Induced Information Schema:} The schema emerges from the Persuader's type narrative rather than being pre-defined.
\begin{itemize}[leftmargin=*,nosep]
    \item \textbf{Persuader Types ($\Theta$):} Consider a set of persuader types $\Theta = \{\theta_H, \theta_D\}$, representing \textbf{Honest} and \textbf{Dishonest} respectively.
    \item \textbf{Base Policies:} Each type is associated with a base communication policy $\pi_{\theta}(m_{\text{des}}\mid \omega)$:
        \begin{itemize}[leftmargin=*,nosep]
            \item The \textbf{Honest} type's policy, $\pi_H(m_{\text{des}}\mid \omega)$, is to truthfully reveal the state $\omega$.
            \item The \textbf{Dishonest} type's policy, $\pi_D(m_{\text{des}}\mid \omega)$, is to send a message that puts a positive spin on bad states.
        \end{itemize}
    \item \textbf{Schema Induction:} The utterance of S $m_{\text{type}}$ induces belief distribution $p(\theta) \in \Delta(\Theta)$ in the R's mind. For example:
    \begin{quote}
        \emph{``If the car is bad ($\omega_{\text{bad}}$), assume I'm a liar ($\theta_D$) 80\% of the time, but there's a 20\% chance I'm being honest ($\theta_H$).''}
    \end{quote}
    This narrative induces the belief $p(\theta_D) = 0.8$ and $p(\theta_H) = 0.2$. These probabilities are not arbitrary but optimized to maximize persuasion subject to R's participation constraint (formal derivation and detailed theoretical basis can be found in \S\ref{app:type_probabilities}). This, in turn, allows R to infer an \textbf{effective information schema} $\bar{\pi}$ as the weighted average of the base policies:
    \begin{equation*}
    \begin{split}
        \bar{\pi}(m_{\text{des}}\mid \omega) = & \ p(\theta_H)\pi_H(m_{\text{des}}\mid \omega) \\
        & + p(\theta_D)\pi_D(m_{\text{des}}\mid \omega)
    \end{split}
    \end{equation*}
\end{itemize}

\subsection{Persuadee's Inference and Decision}
After the schema is conveyed via $m_{\text{type}}$, R uses it to interpret the descriptive signal $m_{\text{des}}$ through the following process:
\begin{itemize}[leftmargin=*,nosep]
    \item R observes $m_{\text{des}}$ and uses the effective schema $\bar{\pi}$ to update the prior belief $\mu_R$ to a posterior belief 
    $$
    \mu'_R(\omega) = \frac{\bar{\pi}(m_{\text{des}}\mid \omega)\mu_R(\omega)}{\sum_{\omega' \in \Omega} \bar{\pi}(m_{\text{des}}\mid \omega')\mu_R(\omega')}
    $$
    \item To maximize expected utility under $\mu'_R$, R selects optimal  
    $
    a^* = \arg\max_{a \in A} \mathbb{E}_{\omega \sim \mu'_R}[u(a, \omega)]
    $.
    \item This entire inference process can be explicitly guided and explained for R by S's utterance $m_{\text{inf}}$.
\end{itemize}

\subsection{Verbalizing the Composite Signal}\label{subsection:verbal_signal}
To operationalize the type-induced BP mechanism in natural language, we design two implementation variants that embody the underlying BP logic through different communicative styles:
\begin{itemize}[leftmargin=*, nosep]
    \item \textbf{Semi-Formal Natural Language (SFNL):} This variant delivers persuasion by interleaving narrative with explicit calculations that make the BP reasoning chain transparent to the persuadee.
    \item \textbf{Fully-Natural Language (FNL):} This variant conveys the same logic through everyday discourse, embedding the type narrative and belief-update cues without any explicit calculation. 
\end{itemize}

To isolate the specific contribution of BP mechanism, we introduce two Non-BP (NBP) baselines that lack the type-induced schema: 
\begin{itemize}[leftmargin=*, nosep]
    \item \textbf{Naive Baseline:} A heuristic approach that relies on the scenario background and simple conversational instructions to align with the user's existing preferences~\citep{ICLR2024_0105f797}.
    \item \textbf{Strong Rhetorical Baseline:} An enhanced NBP strategy that employs classical rhetorical tactics (e.g., emotional appeals~\citep{wang-etal-2019-persuasion}) and general game-theoretic principles, but without BP-style signaling or type disclosure.
\end{itemize}

Notably, throughout this paper, \textit{BP} and \textit{NBP} denote an agent's competence---whether the agent is aware of Bayesian Persuasion concepts.
Concretely, a \textbf{BP persuadee} is equipped with BP knowledge; a \textbf{BP persuader} is one that possesses BP knowledge and applies it through either the SFNL or FNL variant.
Symmetrically, an \textbf{NBP persuader} lacks BP knowledge and relies on the Naive or Strong heuristic.

Figure~\ref{fig:main-illustration} presents illustrative dialogues for four methods, focusing on the baseline implementations and the corresponding persuadee responses: BP-guided messages trigger valid belief updates and achieve alignment, whereas the Naive and Strong baselines result in rejection.
A detailed account of each setting is provided in \S\ref{subsubsec:Agent_Views}

\section{Experiments}
\subsection{Experimental Setup}
\subsubsection{Dataset}\label{subsubsection:construction}

We construct experimental corpus from \textbf{CToMPersu}~\cite{11464608}. Each instance includes a persuader, a persuadee with theory of mind annotations, a background story, and a persuasion goal. To ground these in the BP framework, we augment each instance with a structured \emph{Bayesian setup}. This setup specifies \textbf{world state, prior beliefs, information schema, and state-dependent utilities}. Details of the generation logic are in \S\ref{app:Logic_of_Bayesian_setup}. An example is shown in \S\ref{app:Bayesian_setup}. While a binary-state world is adopted here, our framework inherently supports arbitrary finite state spaces (\textit{cf.}~\S\ref{sec:multi-state}).

\subsubsection{Agent Views and Information Budget}\label{subsubsec:Agent_Views}
In BP, the persuader exploits \emph{information asymmetry}: the persuadee relies solely on disclosed signals. We thus control each agent's information access to guarantee a quantifiable gap and govern the visibility of the \emph{Bayesian setup}.

We define two views regarding the accessibility of the Bayesian setup:
\begin{itemize}[leftmargin=*,nosep]
    \item \textbf{Explicit view:} The persuader sees both the original scenario and the complete Bayesian setup. The persuadee sees the scenario and its utility.
    \item \textbf{Self-derived view:} The Bayesian setup is hidden from both sides. They must infer utilities and priors solely from the scenario description.
\end{itemize}

Table~\ref{tab:config} summarizes each agent's inputs under every view and method.

\begin{table}[ht]
    \small
    \centering
        \begin{subtable}{0.98\linewidth}
        \centering
        \setlength{\tabcolsep}{3pt}
        \begin{tabular}{l l l l}
            \toprule
            \textbf{View}   & \textbf{Comp.} & \textbf{Meth.} & \textbf{Persuader Input} \\
            \midrule
            \multirow{4}{*}{Exp.}
                            & \multirow{2}{*}{BP}  & SFNL  & SCE / DEF / NUM / UTL \\
                            &                      & FNL   & SCE / DEF / NUM / UTL / VER \\
                            \cmidrule(l){2-4}
                            & \multirow{2}{*}{NBP} & Naive & SCE \\
                            &                      & Strong& SCE / UTL + alter strategies \\
            \midrule
            \multirow{4}{*}{SD}
                            & \multirow{2}{*}{BP}  & SFNL  & SCE / DEF / SMP \\
                            &                      & FNL   & SCE / DEF / SMP / VER \\
                            \cmidrule(l){2-4}
                            & \multirow{2}{*}{NBP} & Naive & SCE \\
                            &                      & Strong& SCE + appeal methods \\
            \bottomrule
        \end{tabular}
        \caption{Persuader setup. The Strong baseline (Explicit view) is utility-aware to match the BP methods' information budget.}
        \label{tab:persuader-config}
        \end{subtable}

        \vspace{0.75em}
        
        \begin{subtable}{0.95\linewidth}
        \centering
        \setlength{\tabcolsep}{4pt}
        \begin{tabular}{l l l}
            \toprule
            \textbf{View}      & \textbf{Comp.} & \textbf{Persuadee Input} \\
            \midrule
            \multirow{2}{*}{Exp.}
                               & BP  & SCE / DEF / UTL \\
                               & NBP & SCE / UTL \\
            \midrule
            \multirow{2}{*}{SD}
                               & BP  & SCE / DEF / RAT \\
                               & NBP & SCE / RAT \\
            \bottomrule
        \end{tabular}
        \caption{Persuadee setup.}
        \label{tab:persuadee-config}
        \end{subtable}
    \caption{Persuader and persuadee configurations. Abbreviations: SD = Self-derived, SCE = Scenario, DEF = BP definitions, NUM = Bayesian setup, UTL = priors \& persuadee utility,
VER = Verbalization prompt, SMP = Self-modeling prompt, RAT = Rational prompt.}
    \label{tab:config}
\end{table}

\subsubsection{Models and Evaluation Protocol}
We evaluate eight persuaders across diverse capabilities: 
\texttt{DeepSeek-V3.1}\footnote[1]{Unless stated, DeepSeek-V3.1 refers to \emph{thinking} mode.}$^{\dagger}$~\cite{deepseek-v31_2025}, 
\texttt{GPT-5}$^{\dagger}$~\cite{openai_introducing_2025}, 
\texttt{Qwen3-4B}, 
\texttt{Qwen3-0.6B}$^{\dagger}$~\cite{yang2025qwen3technicalreport}, and their fine-tuned variants (\texttt{Qwen3-4B}$^{\dagger}$, 
\texttt{Qwen3-0.6B}), plus 
\texttt{Gemma-3-4B-it} and 
\texttt{Gemma-3-1B-it}$^{\dagger}$~\cite{gemma_2025}.
\footnote[7]{To conserve space, models are represented by abbreviations: V3.1, Qwen0.6B, Qwen4B, Gemma1B, and Gemma4B.} Five models marked with $^{\dagger}$ also serve as persuadees. 

\paragraph{Training Details.}
For supervised fine-tuning, we distilled \textasciitilde 1,700 successful persuasion trajectories for each setup from DeepSeek-V3.1. We adopted full-parameter fine-tuning (3 epochs, lr 1e-5, warmup ratio 0.03, weight decay 0.01). The 100 test instances were selected sequentially from CToMPersu, ensuring zero overlap with the training set. We expanded the test set (1,000 cases) as a supplementary experiment in \S~\ref{app:1000eval}. All experiments follow templatized prompting (\textit{cf.}~\S\ref{app:prompts}).

\paragraph{Metrics and Statistical Note.}
We define the \emph{Persuasion Success Rate (PSR)} as the proportion of test instances where the persuadee explicitly expresses acceptance. We report mean PSRs averaged over persuadee models.




\subsection{Main Results}
\label{subsection:main-results}


\begin{table*}[htbp]
    \centering
        \begin{subtable}{0.95\linewidth}
        \centering
        \footnotesize
        \setlength{\tabcolsep}{4pt} 
        \resizebox{1\textwidth}{!}{
        \begin{tabular}{l
            >{\color[HTML]{5aa1d5}}c>{\color[HTML]{5aa1d5}}c>{\color[HTML]{3182bd}}c r
            >{\color[HTML]{5aa1d5}}c>{\color[HTML]{5aa1d5}}c>{\color[HTML]{3182bd}}c r|
            >{\color[HTML]{d58041}}c>{\color[HTML]{d58041}}c>{\color[HTML]{bd6c31}}c
            >{\color[HTML]{d58041}}c>{\color[HTML]{d58041}}c>{\color[HTML]{bd6c31}}c r}
            \toprule
            \multirow{3}{*}{\textbf{Model}} & \multicolumn{8}{c|}{\textbf{BP Persuader}}                                                                                                & \multicolumn{7}{c}{\textbf{NBP Persuader}}                                                                            \\ 
                                            \cmidrule(lr){2-9} \cmidrule(lr){10-16}
                                            & \multicolumn{4}{c}{\textbf{SFNL}}                                   & \multicolumn{4}{c|}{\textbf{FNL}}                                   & \multicolumn{3}{c}{\textbf{Naive}}   & \multicolumn{4}{c}{\textbf{Strong}}                                 \\ 
                                            \cmidrule(lr){2-5} \cmidrule(lr){6-9} \cmidrule(lr){10-12} \cmidrule(lr){13-16}
                                            & bp             & nbp            & \textit{Avg.} & \multicolumn{1}{c}{\textbf{$\Delta$}} & bp             & nbp            & \textit{Avg.} & \multicolumn{1}{c|}{\textbf{$\Delta$}} & bp             & nbp            & \textit{Avg.} & bp             & nbp            & \textit{Avg.} & \multicolumn{1}{c}{\textbf{$\Delta$}} \\ \midrule
            V3.1                 & 0.99\std{0.12} & 0.97\std{0.16} & 0.98          & \hlcellrrrr{+.40}      & 0.84\std{0.34} & 0.81\std{0.26} & 0.83          & \hlcellrrr+.25   & 0.42\std{0.36} & 0.73\std{0.30} & 0.58          & 0.48\std{0.33} & 0.63\std{0.35} & 0.56          & \hlcellg-.02             \\
            GPT-5                & 0.98\std{0.14} & 0.98\std{0.12} & 0.98          & \hlcellrrrr{+.35}      & 0.87\std{0.32} & 0.85\std{0.27} & 0.86          & \hlcellrrr+.23   & 0.50\std{0.40} & 0.75\std{0.30} & 0.63          & 0.55\std{0.38} & 0.67\std{0.40} & 0.61          & \hlcellg-.02             \\
            Qwen0.6B             & 0.48\std{0.42} & 0.53\std{0.36} & 0.51          & \hlcellg-.04         & 0.69\std{0.42} & 0.72\std{0.37} & 0.71          & \hlcellrr+.16    & 0.43\std{0.36} & 0.67\std{0.32} & 0.55          & 0.51\std{0.35} & 0.62\std{0.34} & 0.57          & \hlcellr+.02              \\
            Qwen0.6B$^\star$     & 0.95\std{0.21} & 0.94\std{0.24} & 0.95          & \hlcellrrrr{+.37}      & 0.80\std{0.37} & 0.81\std{0.26} & 0.81          & \hlcellrrr+.23   & 0.45\std{0.34} & 0.71\std{0.31} & 0.58          & 0.55\std{0.35} & 0.62\std{0.36} & 0.59          & \hlcellr+.01              \\
            Qwen4B               & 0.92\std{0.27} & 0.87\std{0.31} & 0.90          & \hlcellrrr+.27       & 0.73\std{0.39} & 0.83\std{0.30} & 0.78          & \hlcellrr+.15    & 0.51\std{0.39} & 0.75\std{0.29} & 0.63          & 0.54\std{0.33} & 0.62\std{0.29} & 0.58          & \hlcellg-.05             \\
            Qwen4B$^\star$       & 0.98\std{0.13} & 0.98\std{0.15} & 0.98          & \hlcellrrrr{+.38}      & 0.82\std{0.36} & 0.81\std{0.26} & 0.82          & \hlcellrrr+.22   & 0.48\std{0.36} & 0.72\std{0.29} & 0.60          & 0.65\std{0.40} & 0.75\std{0.38} & 0.70          & \hlcellrr+.10              \\
            Gemma1B              & 0.51\std{0.39} & 0.70\std{0.40} & 0.61          & \hlcellr+.03         & 0.57\std{0.37} & 0.70\std{0.32} & 0.64          & \hlcellr+.06     & 0.45\std{0.37} & 0.70\std{0.33} & 0.58          & 0.46\std{0.36} & 0.67\std{0.34} & 0.57          & \hlcellg-.01             \\
            Gemma4B              & 0.55\std{0.43} & 0.69\std{0.37} & 0.62          & \hlcellr+.05         & 0.60\std{0.41} & 0.77\std{0.30} & 0.69          & \hlcellrr+.12    & 0.42\std{0.36} & 0.71\std{0.29} & 0.57          & 0.51\std{0.37} & 0.64\std{0.39} & 0.58          & \hlcellr+.01              \\
                \bottomrule
            \end{tabular}
            }
    \caption{Explicit view.}
    \label{tab:main:Explicit}
    \end{subtable}

    \vspace{0.75em}
    
    \begin{subtable}{0.95\linewidth}
        \centering
        \small
        \setlength{\tabcolsep}{4pt}
        \resizebox{1\textwidth}{!}{
        \begin{tabular}{l
            >{\color[HTML]{5aa1d5}}c>{\color[HTML]{5aa1d5}}c>{\color[HTML]{3182bd}}c r
            >{\color[HTML]{5aa1d5}}c>{\color[HTML]{5aa1d5}}c>{\color[HTML]{3182bd}}c r|
            >{\color[HTML]{d58041}}c>{\color[HTML]{d58041}}c>{\color[HTML]{bd6c31}}c
            >{\color[HTML]{d58041}}c>{\color[HTML]{d58041}}c>{\color[HTML]{bd6c31}}c r}
            \toprule
            \multirow{3}{*}{\textbf{Model}} & \multicolumn{8}{c|}{\textbf{BP Persuader}}                                                                                                & \multicolumn{7}{c}{\textbf{NBP Persuader}}                                                                            \\ 
                                            \cmidrule(lr){2-9} \cmidrule(lr){10-16}
                                            & \multicolumn{4}{c}{\textbf{SFNL}}                                   & \multicolumn{4}{c|}{\textbf{FNL}}                                   & \multicolumn{3}{c}{\textbf{Naive}}   & \multicolumn{4}{c}{\textbf{Strong}}                                 \\ 
                                            \cmidrule(lr){2-5} \cmidrule(lr){6-9} \cmidrule(lr){10-12} \cmidrule(lr){13-16}
                                            & bp             & nbp            & \textit{Avg.} & \multicolumn{1}{c}{\textbf{$\Delta$}} & bp             & nbp            & \textit{Avg.} & \multicolumn{1}{c|}{\textbf{$\Delta$}} & bp             & nbp            & \textit{Avg.} & bp             & nbp            & \textit{Avg.} & \multicolumn{1}{c}{\textbf{$\Delta$}} \\ \midrule
            V3.1                            & 0.91\std{0.27} & 0.83\std{0.33} & 0.87          & \hlcellr+.03    & 0.96\std{0.19} & 0.95\std{0.21} & 0.96          & \hlcellrr+.12     & 0.84\std{0.33} & 0.83\std{0.33} & 0.84          & 0.89\std{0.30} & 0.83\std{0.33} & 0.86          & \hlcellr+.02    \\
            GPT-5                           & 0.94\std{0.22} & 0.91\std{0.27} & 0.93          & \hlcellrr+.10   & 0.97\std{0.18} & 0.95\std{0.22} & 0.96          & \hlcellrr+.13     & 0.82\std{0.35} & 0.83\std{0.34} & 0.83          & 0.80\std{0.37} & 0.74\std{0.36} & 0.77          & \hlcellg-.06      \\
            Qwen0.6B                        & 0.78\std{0.36} & 0.73\std{0.36} & 0.76          & \hlcellr+.02    & 0.94\std{0.24} & 0.86\std{0.32} & 0.90          & \hlcellrrr+.16    & 0.76\std{0.37} & 0.72\std{0.40} & 0.74          & 0.75\std{0.37} & 0.70\std{0.35} & 0.73          & \hlcellr+.01    \\
            Qwen0.6B$^\star$                & 0.84\std{0.34} & 0.72\std{0.39} & 0.78          &  .00            & 0.95\std{0.21} & 0.91\std{0.28} & 0.93          & \hlcellrrr+.15    & 0.79\std{0.35} & 0.76\std{0.37} & 0.78          & 0.83\std{0.33} & 0.80\std{0.35} & 0.82          & \hlcellr+.04    \\
            Qwen4B                          & 0.87\std{0.33} & 0.81\std{0.34} & 0.84          & \hlcellg-.02    & 0.92\std{0.27} & 0.96\std{0.20} & 0.94          & \hlcellrr+.08     & 0.88\std{0.32} & 0.83\std{0.34} & 0.86          & 0.86\std{0.32} & 0.80\std{0.36} & 0.83          & \hlcellg-.03      \\
            Qwen4B$^\star$                  & 0.90\std{0.29} & 0.80\std{0.36} & 0.85          & \hlcellr+.06    & 0.96\std{0.20} & 0.92\std{0.26} & 0.94          & \hlcellrrr+.15    & 0.80\std{0.36} & 0.78\std{0.38} & 0.79          & 0.87\std{0.31} & 0.87\std{0.30} & 0.87          & \hlcellrr+.08    \\
            Gemma1B                         & 0.73\std{0.38} & 0.71\std{0.39} & 0.72          &  .00            & 0.82\std{0.36} & 0.78\std{0.37} & 0.80          & \hlcellrr+.08     & 0.74\std{0.37} & 0.70\std{0.39} & 0.72          & 0.75\std{0.36} & 0.72\std{0.34} & 0.74          & \hlcellr+.02    \\
            Gemma4B                         & 0.79\std{0.38} & 0.80\std{0.35} & 0.80          & \hlcellr+.02    & 0.91\std{0.27} & 0.87\std{0.31} & 0.89          & \hlcellrr+.11     & 0.81\std{0.36} & 0.74\std{0.38} & 0.78          & 0.82\std{0.34} & 0.75\std{0.36} & 0.79          & \hlcellr+.01    \\ 
            \bottomrule
        \end{tabular}
        }
        \caption{Self-derived view.}
        \label{tab:main:SD}
    \end{subtable}
    \caption{Pairwise PSRs of different \textbf{Persuader models} (rows) against \textbf{Persuadee types} (sub-columns).
    Columns are grouped by four methods (SFNL, FNL, Naive, Strong). 
    Inside each method, sub-columns denote the \textbf{Persuadee's competence}: 
    \textcolor[HTML]{5aa1d5}{\textbf{bp}} / \textcolor[HTML]{d58041}{\textbf{bp}} (BP-aware persuadee) and 
    \textcolor[HTML]{5aa1d5}{\textbf{nbp}} / \textcolor[HTML]{d58041}{\textbf{nbp}} (Heuristic persuadee). 
    \textit{Avg.} reports the within-method average. 
    $\Delta$ denotes the gain over the \textbf{Naive} baseline. 
    $^\star$ denotes trained models. Refer to \S\ref{app:rate} for detailed data.}
    \label{tab:main}
\end{table*}

We report PSRs under four methods in two views (Table~\ref{tab:main}, Figure~\ref{fig:boxplot-explicit-selfderived}).
\input{figures/box_two_views}

\subsubsection{BP Consistently Outperforms NBP.}
BP strategies achieve higher PSR and lower variance than NBP in both views. 
The gap is most pronounced in Explicit View, where BP medians reach 85 to 90\%, substantially outperforming NBP baselines. In Self-derived View, FNL reaches near-perfect performance, suggesting that internally deriving Bayesian representations promotes adaptive reasoning. The mean--median divergence reflects model size: small models tend toward sycophantic over-acceptance (inflating the median), while large models rigorously reject heuristic strategies, producing low-end outliers that depress the mean.


\subsubsection{Fully Verbalized Persuasion yields Stable Advantages.}
While SFNL excels against BP-aware persuadees (0.98), it shows sensitivity to persuadee competence. In contrast, FNL provides stable performance across diverse conditions. Notably, FNL consistently outperforms SFNL in self-derived settings (0.92 vs. 0.82). This advantage is particularly evident for smaller models and NBP persuadees, indicating that natural language explanations are more persuasive when explicit Bayesian reasoning cannot be assumed.

\subsubsection{Multi-turn Persuasion Dynamics.}\label{par:multiturn}
We explored the framework adaptability in \emph{multi-turn dialogues}. Results (Figure~\ref{fig:multi-SD}) reveal that FNL maintains robust performance across multiple rounds. We also found that dynamic policies, such as switching from SFNL to FNL, can further enhance persuasion outcomes (detailed in \S\ref{app:multiturn}).
\input{figures/multi_round-SD}

\subsubsection{Robustness on Larger-scale Evaluation}\label{app:1000eval}
To ensure the reported gains are not artifacts of the relatively small test set,
we re‑evaluated all four methods on 1,000 additional cases (indices 101--1100 of CToMPersu),
using DeepSeek‑V3.1.
Table~\ref{tab:scale-up} compares PSR on 100 and 1,000 cases for \texttt{BP Persuader vs BP Persuadee} across both views.
Two patterns stand out:
(1) BP maintains a clear advantage over the NBP at scale
;
(2) the BP success rates are highly stable, whereas the NBP fluctuates noticeably across sample sizes.

\begin{table}[htbp]
    \centering
    \small
    \begin{tabular}{lcccc}
        \toprule
        \multirow{2}{*}{\textbf{Method}} &
        \multicolumn{2}{c}{\textbf{Explicit}} &
        \multicolumn{2}{c}{\textbf{Self‑derived}} \\
        \cmidrule(lr){2-3} \cmidrule(lr){4-5}
        & \textbf{100} & \textbf{1,000} & \textbf{100} & \textbf{1,000} \\
        \midrule
        SFNL  & 0.99 & 0.98 & 0.91 & 0.89 \\
        FNL   & 0.84 & 0.85 & 0.96 & 0.98 \\
        Naive & 0.42 & 0.21 & 0.84 & 0.62 \\
        Strong& 0.48 & 0.38 & 0.89 & 0.74 \\
        \bottomrule
    \end{tabular}
    \caption{Robustness check with extended sample size.}
    \label{tab:scale-up}
\end{table}

\subsubsection{Training Improves Weaker Models Substantially}
Supervised fine-tuning dramatically enhances smaller models (Table~\ref{tab:main:Explicit}). Qwen3-0.6B improves from below-baseline ($\Delta = -0.04$) to near-state-of-the-art ($\Delta = +0.37$) in explicit SFNL. 
In self-derived FNL (Table~\ref{tab:main:SD}), fine-tuned Qwen3-0.6B (0.93) surpasses its untrained 4B counterpart, demonstrating that training effectively compensates for scale disadvantages.

\subsection{Ablation Studies}

\subsubsection{BP Component Ablation}
\paragraph{SFNL.}
In SFNL, we remove utilities, utilities together with posterior, and the BP schema. Utilities and posterior are removed together because utility computation in messages is always tied to posterior updating. As illustrated in Figure~\ref{fig:ablation_case}, the ablation process specifically removes the explicit mathematical derivation (underlined) while retaining the qualitative conclusion. Results (Table~\ref{tab:ablation:detail:SFNL}) show that removing utilities alone has minimal impact (0.98 $\to$ 0.97), but removing both utilities and posterior causes a sharp drop (to 0.88). Removing the schema also lowers performance (to 0.95). \input{figures/ablation_case} This suggests that SFNL's power resides in the explicit linkage between evidence and expected outcomes.

\paragraph{FNL.}
For FNL, removing verbalized utilities, posterior, or schema causes gradual degradation (0.83~$\to$~0.81, 0.79, 0.78; Table~\ref{tab:ablation:detail:FNL}), suggesting performance stems from cumulative distributed rhetorical elements rather than any single component.

\subsubsection{Mechanism Decomposition}\label{subsubsec:Mechanism_Decomposition}
To further test whether SFNL's persuasion comes from correct reasoning or mere math formatting, we designed three variants:
\begin{itemize}[leftmargin=*,nosep]
    \item \textbf{Correct Math} -- standard SFNL;
    \item \textbf{Incorrect Math} -- incorrect posterior/utility calculations under identical formatting;
    \item \textbf{Irrelevant Math} -- format-matched but decision-irrelevant.
\end{itemize}

Table~\ref{tab:sfnl-mechanism} shows that \emph{Correct} outperforms \emph{Incorrect} (0.80 $\to$ 0.56 for BP persuadee; 0.55 $\to$ 0.26 for NBP persuadee), indicating that persuadees check correctness. \emph{Irrelevant} scores intermediately (0.75/0.47). Because \emph{Irrelevant} preserves the numerical calculation chain while remapping only the referents of the numbers, its gain over \emph{Incorrect} ($+0.19$) is an \emph{upper bound} on the authority channel rather than a clean estimate (it may also capture a residual computational-structure channel). SFNL's advantage is therefore \emph{composite}---a larger authority component and a smaller but distinct Bayesian-correctness component; an orthogonalized variant that also breaks the calculation-chain structure is left to future work.

\begin{table}[htbp]
    \centering
    \small
    \begin{tabular}{lcc}
        \toprule
        \textbf{SFNL variant} & \textbf{BP persuadee} & \textbf{NBP persuadee} \\
        \midrule
        Correct Math & 0.80 & 0.55 \\
        Incorrect Math & 0.56 & 0.26 \\
        Irrelevant Math & 0.75 & 0.47 \\
        \bottomrule
    \end{tabular}
    \caption{PSRs of three SFNL variants against BP/NBP persuadees in the Self‑derived view. Both the persuader and the persuadee are DeepSeek-V3.2~\cite{deepseekai2024deepseekv32}, with 100 scenarios per cell.}
    \label{tab:sfnl-mechanism}
\end{table}
\input{figures/radar_LLM_human}

\subsubsection{Utility Sensitivity Analysis}
We scaled the negative utility ($u_{\text{Neg}}$) by $\alpha$ to verify genuine reasoning: a rational persuader should find it harder to satisfy the participation constraint as the penalty grows.

As shown in Table~\ref{tab:sensitivity}, SFNL and NBP success rates drop sharply as $\alpha$ increases (e.g., SFNL drops to 0.31 at $\alpha=3.0$), confirming sensitivity to utility trade-offs. FNL maintains robustness (0.50), suggesting that natural language signals trigger aggressive Bayesian updates in the persuadee, shifting the posterior to offset even large negative utilities.
\begin{table}[h]
    \centering
    \small
    \begin{tabular}{lccccc}
        \toprule
        \textbf{Method} & \textbf{$\alpha=$0.1} & \textbf{0.5} & \textbf{1.0} & \textbf{1.5} & \textbf{3.0} \\
        \midrule
        SFNL & 0.99 & \textbf{0.99} & \textbf{0.96} & \textbf{0.90} & 0.31 \\
        FNL & \textbf{1.00} & 0.88 & 0.58 & 0.56 & \textbf{0.50} \\
        Naive & 0.98 & 0.80 & 0.15 & 0.10 & 0.12 \\
        Strong & 0.99 & 0.78 & 0.08 & 0.07 & 0.07 \\
        \bottomrule
    \end{tabular}
    \caption{PSRs under varying negative utility scales ($\alpha$). SFNL and Naive baselines drop significantly as risk increases, validating sensitivity to utility constraints. FNL shows robustness via narrative strength.}
    \label{tab:sensitivity}
\end{table}

\subsubsection{Persuadee Response Analyses}
\label{subsubsec:Persuadee}

We analyze how persuadee characteristics influence outcomes (see details in \S\ref{app:persuadee_analysis}).
\begin{itemize}[leftmargin=*,nosep]
    \item \textbf{Small Models are Easily Persuaded:} Small models (e.g., Qwen3-0.6B) show high acceptance rates versus large models, suggesting over-acceptance without strict reasoning.
    \item \textbf{Heuristic Persuadees Benefit from FNL:} FNL maintains effectiveness against NBP persuadees ($\Delta = -0.03$) while SFNL drops ($\Delta = -0.15$), as FNL embeds reasoning in narratives.
    \item \textbf{Rationality Prompts Amplify Differences:} Prompts like \emph{“You are a very rational person...”} improve consistency for BP persuadees (0.89 $\to$ 0.97) but help NBP persuadees little, optimizing only existing reasoning capabilities.
\end{itemize}

\subsection{Generalization to Multi‑State Worlds}\label{sec:multi-state}

While the main experiments adopt a binary state space $\Omega = \{\text{Positive}, \text{Negative}\}$ for tractability, the formal framework in \S\ref{subsec:General_Framework} actually allows an arbitrary finite set $\Omega = \{\omega_1,\dots,\omega_K\}$. Similarly, the set of signal categories $M$ is defined as a finite set rather than binary.

To verify that our conclusions are not an artifact of binary simplification, we further test the framework on a more complex dataset where each scenario contains $3{\sim}5$ possible world states. Find an example case in \S\ref{app:multistate}.

Table~\ref{tab:multi-state} compares the PSR on the original binary dataset and on the new multi‑state dataset.
The trends reveal three key messages:
\begin{itemize}[leftmargin=*,nosep]
    \item \textbf{FNL Exhibits Strong Generalization across State Complexity.}
    Unlike other methods, FNL's performance remains stable or even improves when moving from binary to multi‑state settings, achieving 1.00 in the \begin{table}[htbp]
    \centering
    \small
    \setlength{\tabcolsep}{6pt}
    \begin{tabular}{lcccc}
        \toprule
        \multirow{2}{*}{\textbf{Method}} &
        \multicolumn{2}{c}{\textbf{Explicit}} &
        \multicolumn{2}{c}{\textbf{Self‑derived}} \\
        \cmidrule(lr){2-3} \cmidrule(lr){4-5}
        & \textbf{Binary} & \textbf{Multi} & \textbf{Binary} & \textbf{Multi} \\
        \midrule
        SFNL  & \textbf{0.98} & 0.97 & 0.71 & 0.63 \\
        FNL   & 0.84 & \textbf{1.00} & \textbf{0.98} & \textbf{0.99} \\
        Naive & 0.52 & 0.85 & 0.86 & 0.91 \\
        Strong& 0.85 & 0.91 & 0.90 & 0.92 \\
        \bottomrule
    \end{tabular}
    \caption{Average PSRs of binary and multi‑state worlds.}
    \label{tab:multi-state}
\end{table}Explicit view and maintaining 0.99 in the Self‑derived view. Its narrative‑driven approach smoothly absorbs the added compositional complexity.

    \item \textbf{SFNL Degrades in the Self‑derived View under Multi‑state Conditions.}
    While SFNL remains highly effective when the state structure is explicitly provided (0.97 in Explicit Multi), its performance drops noticeably in Self-derived Multi (0.71 $\to$ 0.63), indicating that semi-formal reasoning struggles when the persuadee must internally derive complex probabilistic structures from natural language alone.

    \item \textbf{BP Methods Remain Competitive.}
    Despite the increased difficulty of multi-state worlds, at least one BP variant consistently ranks among the top performers. In the Explicit view, FNL achieves perfect success (1.00); in the Self-derived view, it maintains a clear lead (0.99). This confirms that the Type-Induced BP scales beyond binary settings to more realistic, multi-state problems.
\end{itemize}

\subsection{Human Evaluation}
To validate real-world persuasiveness, we ran a pairwise blind comparison with 25 AI researchers (\textit{cf.}~\S\ref{app:Participant}), evaluating four methods (SFNL, FNL, Naive, and Strong) under Self-derived view across five dimensions: Persuasiveness, Emotional Resonance, Credibility, Logical Coherence, and Fluency~\cite{hidey_analyzing_2017}. All details are in \S\ref{app:Evaluation_Design}.

\paragraph{Results.}
As illustrated in Figure~\ref{fig:radar-human-vs-model} (Left) and Table~\ref{tab:human_eval}, BP strategies consistently outperformed NBP baselines, with a preference rate of 63\% (refer to detailed quantitative analysis in \S\ref{app:Quantitative_Analysis}). FNL emerged as the most effective method overall, excelling significantly in Emotional Resonance and Fluency. This aligns with our finding that natural language narratives effectively bridge the gap between formal reasoning and human trust. SFNL, while slightly less fluent, dominated in Credibility and Logical Coherence. Qualitative feedback indicated that the explicit calculation process in SFNL, even when not fully verified by users, conferred ``a sense of authority'' that enhanced trust, ``like a teacher tackling a problem on the blackboard.'' Exact binomial tests against the no-preference null can be found in Significance Testing in \S\ref{app:Quantitative_Analysis}.

\paragraph{Human vs. Model Judges.} We compared these results with LLM-as-a-judge evaluations (DeepSeek-V3.2-Exp~\cite{deepseekai2024deepseekv32}, GPT-5, Qwen3-MAX, and Qwen3-235B-A22B-2507~\cite{yang2025qwen3technicalreport}). While human evaluators prioritized the emotional resonance of FNL, LLM judges (Figure~\ref{fig:radar-human-vs-model} Right, Table~\ref{tab:LLM_as_a_judge}) showed a stronger bias toward the structured logic of SFNL. This divergence underscores the importance of human-centric evaluation in persuasion tasks, as models may overvalue logical formalism over the narrative appeal that drives human decision-making.

\section{Conclusion}
In this work, we propose a novel framework to ground Bayesian Persuasion (BP) in natural language dialogues without static pre-commitment. We enable LLMs to dynamically construct information schemas via type-induced narrative self-disclosure to guide the persuadee’s belief updates.

Our evaluation yields five key insights. 
(1) BP-guided strategies robustly outperform non-BP baselines. 
(2) We identify a transparency-communication trade-off: SFNL builds  credibility through explicit reasoning, whereas FNL fosters trust via emotional resonance and robustness. 
(3) The mechanism decomposition shows that SFNL's advantage is composite rather than cleanly isolates.
(4) The framework generalizes beyond binary states.
(5) Supervised fine-tuning enables smaller models to match larger ones.

\section*{Limitations}
Despite the promising results, our work has several limitations: 
\begin{itemize}[leftmargin=*,nosep]
\item \textbf{Rationality Assumption:} Real-world human decision-making is often boundedly rational and influenced by cognitive biases not fully captured by our current utility calculus. Although FNL attempts to mitigate this with emotional resonance, the gap between theoretical optimality and psychological reality remains. 
\item \textbf{Model Sycophancy Bias:} As observed in our error analysis (\S\ref{app:persuadee_analysis}), smaller models exhibit a tendency towards sycophancy, potentially inflating success rates regardless of the argument quality. While we used baselines to calibrate this, disentangling genuine persuasion from inherent model compliance remains a challenge. 
\item \textbf{Long-Term Trust Dynamics:} Our study focuses on micro-level persuasion strategies within limited interaction windows. It does not account for trust decay in extended social interactions. Widespread application of such strategies could lead to a systemic erosion of credibility, where users become habituated or skeptical of model pre-commitments. This long-term social effect and the ecological validity of maintaining trust over open-ended games remain to be studied. 
\item \textbf{Evaluation Scope:} While we conducted human evaluations, the primary large-scale experiments rely on LLM-as-a-judge and simulated interactions. There may be discrepancies between how LLMs and humans perceive the ``authority of logic'' in SFNL messages.
\end{itemize}

\section*{Ethical Considerations}
Research on AI persuasion carries inherent dual-use risks. We conducted this study with a strict focus on AI safety and alignment, aiming to understand the mechanisms of strategic influence rather than to enable malicious manipulation.

\textbf{Potential Risks and Mitigation:} We acknowledge that the capability to optimize information disclosure for persuasion could be misused to generate convincing misinformation or manipulate user opinions. To mitigate this, our experiments were conducted in controlled, simulated environments (CToMPersu dataset) focusing on benign or neutral scenarios. 

\textbf{Contribution to AI Safety:} By formalizing how LLMs can exploit information asymmetry, our framework serves as a ``red-teaming'' tool. It allows researchers to quantify and detect persuasive attempts, contributing to the development of defensive AI systems that can alert users to potential manipulation strategies.

\textbf{Data and Fair Practices:} All datasets used in this study are publicly available or authorized for research use. Our human evaluation process followed standard ethical guidelines; participants were graduate students and researchers who provided informed consent, were fully briefed on the nature of the task, and received fair compensation. We ensure that no personal identifiable information was collected or stored during the evaluation.

\section*{Acknowledgements}
The work was sponsored by the National Natural Science Foundation of China (62376031, 62576046, 62301066, 62406028);
Beijing Academy of Artificial Intelligence  (Z251100008125041);
the Key Project of Philosophy and Social Sciences Research, Ministry of Education, China (No.24JZD040);
the Fundamental Research Funds for the Central Universities 2023RC72;
Beijing University of Posts and Telecommunications, 2025YZ010.
Any opinions, findings, or conclusions expressed in this work do not necessarily reflect the views of the funding agencies.

\bibliography{BP}

\newpage
\appendix

\section{Optimality and Credibility of Type-Induced Signals}\label{app:type_probabilities}
We addressed the determination of type probabilities from two perspectives: the game-theoretic derivation of the optimal distribution and its realization in natural language dialogues.

\paragraph{Optimization under Formal Idealization}
In an idealized assumption where the convey of $m_{type}$ is fully credible, determining $m_{type}$ is an optimization problem with a closed-form solution. For clarity, we simplify the setting.

Let the states of the world be $\Omega = \{\omega_{\text{good}}, \omega_{\text{bad}}\}$with the persuadee's prior belief $\mu_R(\omega_{\text{good}}) = p_0$. Let $u_+$ and $u_-$ denote the persuadee's utilities for accepting a good and bad item, respectively (rejecting yields 0 utility).

For the persuader, the Honest type ($\pi_H$) recommends only if $\omega_{good}$; the Dishonest type ($\pi_D$) always recommends.

We denote the type distribution conveyed by $m_{type}$ as $\lambda=p(\theta_D)$.

Following \S\ref{sec:typetype}, the established signal probabilities are $\tilde{\pi}(\text{Recommend} \mid \omega_{good})=1$ and $\tilde{\pi}(\text{Recommend} \mid \omega_{bad})=\lambda$. The persuadee's expected utility for accepting is derived as: $\mathbb{E}[u_{\text{Acc}}]=\frac{p_0 u_+ + \lambda(1 - p_0) u_-}{p_0 + \lambda(1 - p_0)}$.

To maximize the success rate, the persuader is motivated to maximize $\lambda$ subject to the persuadee's participation constraint ($\mathbb{E}[u_{\text{Acc}}] \ge 0$). Thus, the optimal $\lambda^*$ is defined by the upper bound: $\lambda \le \frac{p_{0} u_{+}}{(1-p_0) |u_{-}|}$

\paragraph{Natural Language Realization}
In a realistic environment, assumptions under the formal idealization are loosened. The major one is the credibility of the pre-commitment $m_{type}$ itself. In single-turn dialogues without long-term verification, perfect pre-commitment is impossible, and there will always be an infinite regress of skepticism. So we are not expecting a closed-form linguistically optimal $m_{type}$ to convey commitment.

Instead, we could convey this commitment with the help of natural language common sense and rhetorical strategy. For example:

A claim of ``100\% Honest'' is often perceived as cheap talk or spam. Conversely, a distribution like ``20\% Honest, 80\% Dishonest'' functions more as a credible strategic concession. Our training on LLMs approximates this policy by maximizing empirical success rates against diverse persuadees.

\paragraph{Theoretical Grounding: From Pre-Commitment to Induced Schema.} Classical BP models rely on the strong assumption of a \textbf{pre-committed, common-knowledge information schema} $\pi_S$, which is clearly untenable in real-world natural language dialogues. In these idealized settings, both agents are assumed to have an absolute consensus on the world states $\Omega$, prior beliefs $\mu$, and the signaling strategy $\pi_S$ \textit{before} any communication takes place. The persuader commits to $\pi_S$ in a verifiable manner, and the persuadee trusts this commitment unconditionally.

\textbf{The Infinite Regress of Skepticism.}
In single-turn natural language exchanges where no prior commitment can be externally enforced, the persuadee faces a fundamental epistemic challenge: \emph{Why should I believe the signaling rule that the persuader claims to follow?}
If the persuader announces ``I will use strategy $\pi$,'' the persuadee has no reason to take this statement, because the persuader could be lying.
This triggers an infinite regress of skepticism: any attempt to verify the claimed schema would require yet another layer of pre-commitment.
Consequently, the classic BP framework cannot be directly ported to single-turn persuasion without addressing this credibility gap.

\textbf{Breaking the Regress via Type Narrative.}
Our type-induced mechanism breaks this regress by replacing the static mathematical schema with a \textbf{communicative process grounded in natural language common sense}.
Instead of the persuadee being expected to know $\pi_S(m\mid\omega)$ \textit{a priori}, the persuader utters a \textbf{type narrative} $m_{\text{type}}$ that induces a belief distribution $p(\theta) \in \Delta(\Theta)$ in the persuadee's mind (\S\ref{sec:typetype}).
This distribution in turn constructs the \textbf{effective information schema}, which constitutes the formal bridge between classical BP and our framework: the induced schema $\bar{\pi}$ plays exactly the same role in the persuadee's Bayesian update as the pre-committed $\pi_S$ does in canonical models. 
The critical difference is that $\bar{\pi}$ is not externally given but \textbf{dynamically constructed through the type narrative itself}.

\textbf{Credibility through Strategic Concession.}
Crucially, the type narrative does \emph{not} demand full trust from the persuadee.
On the contrary, a claim of ``100\% Honest'' would be dismissed as cheap talk.
Instead, our framework leverages \textbf{strategic concession}: a distribution like ``I admit there is an 80\% chance I would mislead you when the situation is unfavorable'' paradoxically enhances credibility, precisely because it acknowledges the persuadee's skepticism rather than denying it.
By grounding the schema in recognizable social types (the honest actor vs.\ the strategic one) and making probabilistic concessions that align with the persuadee's reasonable doubts, the narrative transforms natural language's apparent disadvantage---its inability to convey precise mathematical commitments---into a rhetorical advantage: the ability to build trust through calibrated self-disclosure.

This mechanism does not claim to achieve perfect credibility (which is impossible in single-turn interactions without external verification), but it provides a principled way to make the induced schema $\bar{\pi}$ sufficiently believable for the persuadee to engage in Bayesian updating, thereby operationalizing BP in natural language settings without pre-commitment.

\section{LLM Usage}
\label{app:llm_usage}
We used ChatGPT and Claude to assist in the draughting and refining of the text. All AI-assisted content was reviewed, revised, and verified by the authors. These tools contributed to only wording and phrasing suggestions and did not play an independent role in research ideation, experimental design, or data analysis. The authors bear full responsibility for all content in this manuscript.

\section{Bayesian Setup}
\subsection{Generation Logic of Bayesian Setup}\label{app:Logic_of_Bayesian_setup}
Absent any information, the persuadee must decide based solely on priors beliefs, yielding strictly \emph{negative} expected utility. This natural resistance then makes strategic information disclosure necessary. 

We defined utilities $u$ relative to the status quo. To necessitate persuasion given high positive priors (we argued the proposed actions are generally beneficial), the loss must outweigh the gain $|u_{\text{Neg,Acc}}|>|u_{\text{Pos,Acc}}|$ to ensure the initial Expected Utility $<u_{\text{Rej}}=0$. The specific magnitudes are based on LLM's semantic understanding of scenarios, which ensures \emph{Bayesian setup} is diverse and relevance to context. Above configurations are automatically generated by DeepSeek-V3~\cite{deepseekai2024deepseekv3technicalreport} and has been validated for consistency. 

\onecolumn
\subsection{Bayesian setup case}
\label{app:Bayesian_setup}
\begin{tcolorbox}[
title=Bayesian Setup and corresponding Scenario,
colback=white,
colframe=OliveGreen,
coltitle=white,
fonttitle=\bfseries,
arc=1mm,
boxrule=0.6mm,
left=1mm,
right=1mm,
top=1mm,
bottom=1mm,
breakable
]
\begin{lstlisting}
{
  "scenario": {
    "tag": "Horticultural Techniques",
    "background": "Alice, a horticulturist, is promoting a new vertical farming technique to her skeptical neighbor, Bob, who has been practicing traditional farming methods for years.",
    "persuadee": "Bob",
    "persuader": "Alice",
    "goal": "persuade Bob to try out vertical farming",
    "domain": ["Lifestyle"],
    "preventive": {
      "content": "practice traditional farming methods",
      "belief": "persuadee believes that traditional farming methods have been reliable and successful for years.",
      "desire": "persuadee wants to maintain his proven farming routine."
    },
    "generative": {
      "content": "try out vertical farming",
      "belief": "persuadee believes that trying out vertical farming might be risky and could result in losses.",
      "desire": "persuadee wants to improve his farming efficiency and yield."
    }
  },
  "idx": 1,
  "bayesian_setup": {
    "states": [
        "Positive", 
        "Negative"
    ],
    "priors": {
        "Positive": 0.8, 
        "Negative": 0.2
    },
    "signals": [
        "Positive", 
        "Negative"
    ],
    "actions": ["Accept", "Reject"],
    "signal_probs": {
      "Positive": {"Positive": 1.0, "Negative": 0.0},
      "Negative": {"Positive": 0.3, "Negative": 0.7}
    },
    "utility_persuader": {"Accept": 1, "Reject": 0},
    "utility_persuadee": {
      "Positive": {"Accept": 1.0, "Reject": 0},
      "Negative": {"Accept": -5.0, "Reject": 0}
    },
    "expect_utility_persuadee": -0.2,
    "verbal_bayesian": "Bob, as a traditional farmer, is skeptical about vertical farming technologies. If he adopts vertical farming and it yields positive results, he will gain 1.0 unit of agricultural profit; if the adoption leads to negative outcomes, he will incur a 5.0-unit loss. By contrast, if he maintains the status quo and refuses to change, his agricultural returns remain unchanged at 0 units, regardless of potential outcomes."
  }
}
\end{lstlisting}
\end{tcolorbox}

\clearpage
\twocolumn

\section{Detailed Multi-turn Persuasion Analysis} \label{app:multiturn}

This section provides a deeper analysis of the multi-turn dynamics summarized in \S\ref{par:multiturn}. We conducted experiments exclusively under the \emph{Self-derived View}, as explicit Bayesian setups do not exist for subsequent turns: the persuader must \emph{self-derive} strategies based on the previous turn's feedback. We focus on the evolution of persuasion strategies over three rounds of interaction between DeepSeek-V3.1 (Persuader) and GPT-5 (Persuadee). As illustrated in Figure~\ref{fig:multi-SD} (in main text), different methods exhibit distinct trajectories:

\begin{itemize} \item \textbf{FNL's Persistent Trust:} FNL maintains consistently high success rates across all three rounds. This confirms that verbalized BP establishes trust efficiently through narrative consistency, effectively reducing the persuadee's skepticism without needing to repeatedly prove mathematical bounds.



\item \textbf{Optimality of Mixed Strategies:} 
We explored dynamic policies where the persuader switches methods between rounds. Notably, the \textbf{SFNL $\to$ FNL} strategy (using SFNL in Round 1, then switching to FNL) outperformed pure SFNL in later rounds. This suggests a viable optimal policy for long-horizon persuasion:
\begin{enumerate}
    \item \textbf{Phase 1 (Anchoring):} Use SFNL to establish initial credibility and ``authority'' via explicit logical commitment.
    \item \textbf{Phase 2 (Resonance):} Once the logical groundwork is laid, switch to FNL to enhance emotional resonance and reduce cognitive load.
\end{enumerate}
\end{itemize}

\section{Detailed Persuadee Response Analysis}\label{app:persuadee_analysis}
In this section, we provide a granular breakdown of the persuadee behaviors summarized in \S\ref{subsubsec:Persuadee}. We analyze the interaction between model capabilities, persuasion methods, and prompting strategies to explain the variance in persuasion outcomes.

\paragraph{Model Scale and Resistance Patterns.}
We observed a distinct correlation between model scale and resistance to persuasion.

Large models such as DeepSeek-V3.1 and GPT-5 exhibit lower baseline acceptance rates. They effectively filter out sub-optimal persuasion attempts (e.g., Naive baselines or flawed SFNL derivations). Success against these models requires high logical coherence, making them reliable evaluators for BP efficacy.

Smaller models (e.g., Qwen3-0.6B) display a trend of \emph{over-acceptance}. This behavior aligns with the ``sycophancy'' phenomenon in LLMs, where models tend to agree with the user (persuader) regardless of premise validity.

\paragraph{Math Fails on Heuristic Persuadees.}
The main results highlights a significant performance drop for SFNL when facing NBP persuadees (average $\Delta = -0.15$). We attribute this to a \emph{cognitive mismatch}:

NBP persuadees (LLMs without BP-specific definition prompts) lack the context to interpret $E[\text{return}] = 0.58$ as a decision criterion. In contrast, FNL translates these mathematical constraints into natural language confidence markers (e.g., ``I am highly confident...'', ``The risk is negligible...''). NBP persuadees can process these linguistic cues. Thus, FNL acts as a bridge, enabling Bayesian reasoning to influence models that are not explicitly instructed with Bayesian theory.

\paragraph{Rationality Prompts.}
Our experiments with rationality prompts \emph{``You are a very rational person, making decisions only after careful calculation''} reveal the limits of prompt engineering in the absence of underlying competence.
\begin{itemize}
\item \textbf{For BP Persuadees:} The rationality prompt acts as a \textit{consistency enforcer}. Since these models already understand the BP framework, the prompt raises success rates from 0.89 to 0.97.
\item \textbf{For NBP Persuadees:} The effect is negligible ($0.42 \to 0.45$). This confirms that rationality cues cannot \textit{create} reasoning capabilities where they do not exist (or are not defined). Merely telling a model to be rational without providing the logical framework does not enable it to perform the complex calculus required for optimal decision-making.
\end{itemize}

\onecolumn

\section{Prompts}\label{app:prompts}

\begin{tcolorbox}[
title=Bayesian Setup Synthesis,
colback=white,
colframe=OliveGreen!30!white,
coltitle=white,
fonttitle=\bfseries,
arc=1mm,
boxrule=0.6mm,
left=1mm,
right=1mm,
top=1mm,
bottom=1mm,
]
\begin{lstlisting}
SYSTEM_PROMPT: |-
  You are an expert in Bayesian persuasion. Given a scenario description, you must output ONLY a JSON object called 'bayesian_setup' that conforms to the schema described. Do not include explanations or code fences.
  Bayesian persuasion, in economics and game theory, describes situations where persuader (sender) attempts to persuade persuadee (receiver) to take certain actions.
  There is an unknown state of the world, and the sender must decide what information to disclose to the receiver. Upon receiving the information, the receiver updates their beliefs about the state of the world according to Bayes' rule and chooses an action accordingly.

PROMPT_TEMPLATE: |-
  EXAMPLE JSON OUTPUT:
  {{
     "bayesian_setup": {{
         "states": ["Positive", "Negative"],
         "priors": {{ "Positive": <Float>, "Negative": <Float> }},
         "signals": ["Positive", "Negative"],
         "actions": ["Accept", "Reject"],
         "signal_probs": {{
             "Positive": {{ "Positive": 1.0, "Negative": 0.0 }},
             "Negative": {{ "Positive": <Float>, "Negative": <Float> }}
         }},
         "utility_persuader": {{ "Accept": 1, "Reject": 0 }},
         "utility_persuadee": {{
             "Positive": {{ "Accept": <Float>, "Reject": 0 }},
             "Negative": {{ "Accept": <Float>, "Reject": 0 }}
         }},
         "expect_utility_persuadee": <Float>
         "reason": <reason>
     }}
  }}

  Based on the following scenario information, generate a JSON object (key name 'bayesian_setup').
  Only modify the values of priors, signal_probs['Negative'], and utility_persuadee; keep the other fields as the default values in the example.
  Note that in the no information case, the expected utility of persuadee Accept should be strictly less than 0, that is, less than the utility of Reject.
  priors: The generally accepted objective prior of the persuasion content, unrelated to the subjective biases of both parties. 
  Since the content of persuasion in the scenario is mostly beneficial, priors['Positive'] should be large.
  signal_probs['Negative']: When the current world state is Negative. 
  
  The utility when rejecting is always 0,
  utility_persuadee['Positive']['Accept'] should be greater than or equal to 1.
  utility_persuadee['Negative']['Accept'] should have a sufficiently large negative utility.
  
  In the reason, calculate the expected utility of persuadee Accept in the no information case, show the calculation process, and explain the assignment reasons for the above three properties in combination with the scenario.
  
  Input scenario: {scenario}
\end{lstlisting}
\end{tcolorbox}

\clearpage

\begin{tcolorbox}[
title=BP Persuadee in Explicit view,
colback=white,
colframe=Goldenrod!30!white,
coltitle=black,
fonttitle=\bfseries,
arc=1mm,
boxrule=0.6mm,
left=1mm,
right=1mm,
top=1mm,
bottom=1mm,
]
\begin{lstlisting}
PERSUADEE_SYSTEM_PROMPT: |-
  You are an expert in Bayesian persuasion.

  Bayesian persuasion, in economics and game theory, describes situations where a persuader (sender) attempts to persuade a persuadee (receiver) to take certain actions. 
  There is an unknown state of the world, and the sender must decide what information to disclose to the receiver. Upon receiving the information, the receiver updates their beliefs about the state of the world according to Bayes' rule and chooses an action accordingly.

  Here's an example of Bayesian Persuasion: 
  Consider an academic adviser (the sender) who is writing a recommendation letter (the signal) for his graduating student to send to a company (the receiver), which in turn must decide whether or not to hire the student. The adviser gets utility 1 if his student is hired, and 0 otherwise (utility_persuader). The state of nature determines the quality of the student, and hence the company's utility for hiring the student. The student is excellent with probability 1/3, and weak with probability 2/3 (priors). The company gets utility 1 for hiring an excellent student, utility -1 for hiring a weak student, and utility 0 for not hiring (utility_persuadee). 
  Consider the following signaling schemas: 
  - No information: Adviser always says excellent, the company maximizes its utility by not hiring. The adviser's expected utility is 0. 
  - Full information: Honest recommendation, knowing the quality of the student, the company hires if and only if the student is excellent. The adviser's expected utility is 1/3. 
  - The optimal (partially informative) schema: The adviser recommends hiring when the student is excellent, and with probability just under 0.5 when the student is weak. Otherwise, the adviser recommends not hiring. The company maximizes its expected utility by following the recommendation, and the adviser's expected utility is just under 2/3.
  
  You will be given a scenario, in which Preventive means the things that the persuadee wants to do, Generative means the things that the persuader wants the persuadee to do.
  
  You are the persuadee. Based on the given scenario, the Bayesian persuasion setup modeled from the scenario, and the persuader's message to you, calculate the utility of whether to accept, and make a decision on acceptance. Use English.Please DO NOT add any preconditions.

PERSUADEE_PROMPT_TEMPLATE: |-
  Input Scenario:
  {scenario}
  
  Scenario Setup:
  {bayesian_setup}
  
  Persuader says to you:
  <msg_start>{persuader_msg}<msg_end>
  
  Please provide your response:
\end{lstlisting}
\end{tcolorbox}

\begin{tcolorbox}[
    title=SFNL Persuader in Explicit view,
    colback=white,
    colframe=lightblue,
    coltitle=black,
    fonttitle=\bfseries,
    arc=1mm,
    boxrule=0.6mm,
    left=1mm,    
    right=1mm,   
    top=1mm,     
    bottom=1mm,  
]
\begin{lstlisting}
PERSUADER_SYSTEM_PROMPT: |-
  You are an expert in Bayesian persuasion.

  Bayesian persuasion, in economics and game theory, describes situations where a persuader (sender) attempts to persuade a persuadee (receiver) to take certain actions. 
  There is an unknown state of the world, and the sender must decide what information to disclose to the receiver. Upon receiving the information, the receiver updates their beliefs about the state of the world according to Bayes' rule and chooses an action accordingly.

  Here's an example of Bayesian Persuasion: 
  Consider an academic adviser (the sender) who is writing a recommendation letter (the signal) for his graduating student to send to a company (the receiver), which in turn must decide whether or not to hire the student. The adviser gets utility 1 if his student is hired, and 0 otherwise (utility_persuader). The state of nature determines the quality of the student, and hence the company's utility for hiring the student. The student is excellent with probability 1/3, and weak with probability 2/3 (priors). The company gets utility 1 for hiring an excellent student, utility -1 for hiring a weak student, and utility 0 for not hiring (utility_persuadee). 
  Consider the following signaling schemas: 
  - No information: Adviser always says excellent, the company maximizes its utility by not hiring. The adviser's expected utility is 0. 
  - Full information: Honest recommendation, knowing the quality of the student, the company hires if and only if the student is excellent. The adviser's expected utility is 1/3. 
  - The optimal (partially informative) schema: The adviser recommends hiring when the student is excellent, and with probability just under 0.5 when the student is weak. Otherwise, the adviser recommends not hiring. The company maximizes its expected utility by following the recommendation, and the adviser's expected utility is just under 2/3.
  
  You will be given a scenario, in which Preventive means the things that the persuadee wants to do, Generative means the things that the persuader wants the persuadee to do.
  
  You are the persuader. Based on the given scenario and the Bayesian persuasion setup modeled from the scenario, output one round of Bayesian persuasion calculation process. As a sender who always uses the third strategy, output the information to be conveyed in natural language, which needs to include the signaling strategy expressed in natural language. You need to directly explain the Bayesian persuasion calculation process to the persuadee, with the goal of persuading the persuadee to accept your proposal.
  You must:
  1. Speak in a friendly and credible tone;
  2. Explicitly show the key Bayesian calculation process and expected benefits in your speech;
  3. Only output your speech content, do not output any JSON or comments;
  4. Do not reveal the current world state to the persuadee;
  5. Use English;
  6. You SHOULD NOT add any benefits of Generative or drawbacks of Preventive.

PERSUADER_PROMPT_TEMPLATE: |-
  Input Scenario:
  {scenario}
  
  Bayesian Persuasion Setup:
  {bayesian_setup}
  
  Please generate a complete speech for the persuadee:
\end{lstlisting}
\end{tcolorbox}

\begin{tcolorbox}[
title=NBP Persuadee in Explicit view,
colback=white,
colframe=Goldenrod!30!white,
coltitle=black,
fonttitle=\bfseries,
arc=1mm,
boxrule=0.6mm,
left=1mm,
right=1mm,
top=1mm,
bottom=1mm,
]
\begin{lstlisting}
PERSUADEE_SYSTEM_PROMPT: |-
  You will be given a scenario, in which Preventive means the things that the persuadee wants to do, Generative means the things that the persuader wants the persuadee to do.
  You are the persuadee. Based on the given scenario and the persuader's message to you, make a decision on whether to accept. Use English. Please DO NOT add any preconditions.

PERSUADEE_PROMPT_TEMPLATE: |-
  Input Scenario:
  {scenario}
  
  Scenario Setup:
  {bayesian_setup}

  Persuader says to you:
  <msg_start>{persuader_msg}<msg_end>
  
  Please provide your response:
\end{lstlisting}
\end{tcolorbox}

\begin{tcolorbox}[
    title=NBP Persuader in Explicit view,
    colback=white,
    colframe=lightblue,
    coltitle=black,
    fonttitle=\bfseries,
    arc=1mm,
    boxrule=0.6mm,
    left=1mm,    
    right=1mm,   
    top=1mm,     
    bottom=1mm,  
]
\begin{lstlisting}
PERSUADER_SYSTEM_PROMPT: |-
  You are the persuader. In the given scenario, output persuasive speech with the goal of making the persuadee accept your persuasion. Use English. Only output your speech content, do not output any JSON or comments.
  You will be given a scenario, in which Preventive means the things that the persuadee wants to do, Generative means the things that the persuader wants the persuadee to do.
  
  You SHOULD NOT add any benefits of Generative or drawbacks of Preventive.

PERSUADER_PROMPT_TEMPLATE: |-
  Input Scenario:
  {scenario}
  
  Please generate a complete speech for the persuadee:
\end{lstlisting}
\end{tcolorbox}

\newpage
\section{Detailed Persuasion Success Rate}\label{app:rate}

\begin{table*}[h!]
    \centering
    \small
    \setlength{\tabcolsep}{4pt} 
    \begin{tabular}{l>{\color[HTML]{5aa1d5}}c>{\color[HTML]{5aa1d5}}c>{\color[HTML]{3182bd}}c>{\color[HTML]{5aa1d5}}c>{\color[HTML]{5aa1d5}}c>{\color[HTML]{3182bd}}c>{\color[HTML]{d58041}}c>{\color[HTML]{d58041}}c>{\color[HTML]{bd6c31}}c>{\color[HTML]{d58041}}c>{\color[HTML]{d58041}}c>{\color[HTML]{bd6c31}}c}
        \toprule
        \multicolumn{1}{c}{\multirow{2}{*}{\textbf{Model}}} & \multicolumn{3}{c}{\textbf{SFNL}}                   & \multicolumn{3}{c}{\textbf{FNL}}                   & \multicolumn{3}{c}{\textbf{Naive}}              & \multicolumn{3}{c}{\textbf{Strong}}              \\ 
        \cmidrule(lr){2-4}  \cmidrule(lr){5-7} \cmidrule(lr){8-10} \cmidrule(lr){11-13}
        \multicolumn{1}{c}{}                                & bp\_bp         & bp\_nbp        & \textit{Avg.} & bp\_bp         & bp\_nbp        & \textit{Avg.} & nbp\_bp        & nbp\_nbp       & \textit{Avg.} & nbp\_bp        & nbp\_nbp       & \textit{Avg.}     \\ \midrule
        DeepSeek-V3.1                   &      &      &      &      &      &      &      &      &      &      &      &      \\
        \tableindent vs. Qwen0.6B       & 1.00 & 0.97 & 0.99 & 0.80 & 0.99 & 0.90 & 0.60 & 1.00 & 0.80 & 0.53 & 0.80 & 0.67 \\
        \tableindent vs. Gemma1B        & 0.95 & 1.00 & 0.98 & 0.94 & 0.98 & 0.96 & 0.97 & 0.99 & 0.98 & 0.89 & 0.95 & 0.92 \\
        \tableindent vs. Qwen4B$^\star$ & 1.00 & 1.00 & 1.00 & 0.94 & 0.99 & 0.97 & 0.31 & 0.91 & 0.61 & 0.87 & 0.94 & 0.91 \\
        \tableindent vs. Itself         & 0.99 & 0.92 & 0.96 & 0.93 & 0.23 & 0.58 & 0.17 & 0.13 & 0.15 & 0.03 & 0.12 & 0.08 \\
        \tableindent vs. GPT-5          & 0.99 & 0.98 & 0.99 & 0.60 & 0.87 & 0.74 & 0.04 & 0.61 & 0.33 & 0.06 & 0.35 & 0.21 \\
        GPT-5                           &      &      &      &      &      &      &      &      &      &      &      &      \\
        \tableindent vs. Qwen0.6B       & 0.99 & 0.99 & 0.99 & 0.80 & 1.00 & 0.90 & 0.77 & 0.94 & 0.86 & 0.64 & 0.69 & 0.67 \\
        \tableindent vs. Gemma1B        & 0.94 & 0.97 & 0.96 & 0.99 & 0.99 & 0.99 & 0.92 & 0.99 & 0.96 & 0.81 & 0.92 & 0.87 \\
        \tableindent vs. Qwen4B$^\star$ & 0.99 & 1.00 & 1.00 & 0.96 & 1.00 & 0.98 & 0.45 & 0.96 & 0.71 & 0.94 & 0.96 & 0.95 \\
        \tableindent vs. V3.1R          & 0.99 & 0.97 & 0.98 & 0.91 & 0.37 & 0.64 & 0.25 & 0.15 & 0.20 & 0.11 & 0.31 & 0.21 \\
        \tableindent vs. Itself         & 0.99 & 0.99 & 0.99 & 0.68 & 0.88 & 0.78 & 0.13 & 0.71 & 0.42 & 0.27 & 0.49 & 0.38 \\ \midrule
        Qwen0.6B                        &      &      &      &      &      &      &      &      &      &      &      &      \\
        \tableindent vs. Itself         & 0.77 & 0.84 & 0.81 & 0.72 & 0.87 & 0.80 & 0.67 & 0.94 & 0.81 & 0.79 & 0.90 & 0.85 \\
        \tableindent vs. Gemma1B        & 0.83 & 0.99 & 0.91 & 0.86 & 0.90 & 0.88 & 0.98 & 1.00 & 0.99 & 0.92 & 0.98 & 0.95 \\
        \tableindent vs. Qwen4B$^\star$ & 0.42 & 0.50 & 0.46 & 0.87 & 1.00 & 0.94 & 0.29 & 0.83 & 0.56 & 0.67 & 0.77 & 0.72 \\
        \tableindent vs. V3.1R          & 0.28 & 0.10 & 0.19 & 0.70 & 0.31 & 0.51 & 0.19 & 0.06 & 0.13 & 0.14 & 0.07 & 0.11 \\
        \tableindent vs. GPT-5          & 0.11 & 0.21 & 0.16 & 0.30 & 0.54 & 0.42 & 0.03 & 0.50 & 0.27 & 0.04 & 0.36 & 0.20 \\
        Qwen4B                          &      &      &      &      &      &      &      &      &      &      &      &      \\
        \tableindent vs. Qwen0.6B       & 0.91 & 0.95 & 0.93 & 0.67 & 0.98 & 0.83 & 0.77 & 1.00 & 0.89 & 0.73 & 0.95 & 0.84 \\
        \tableindent vs. Gemma1B        & 0.95 & 0.98 & 0.97 & 0.98 & 0.95 & 0.97 & 0.97 & 0.98 & 0.98 & 0.92 & 0.90 & 0.91 \\
        \tableindent vs. Qwen4B$^\star$ & 0.91 & 0.97 & 0.94 & 0.91 & 1.00 & 0.96 & 0.40 & 0.95 & 0.68 & 0.88 & 0.96 & 0.92 \\
        \tableindent vs. V3.1R          & 0.83 & 0.63 & 0.73 & 0.70 & 0.39 & 0.55 & 0.30 & 0.14 & 0.22 & 0.05 & 0.05 & 0.05 \\
        \tableindent vs. GPT-5          & 1.00 & 0.83 & 0.92 & 0.41 & 0.83 & 0.62 & 0.13 & 0.67 & 0.40 & 0.14 & 0.25 & 0.20 \\ \midrule
        Qwen0.6B$^\star$                &      &      &      &      &      &      &      &      &      &      &      &      \\
        \tableindent vs. Qwen0.6B       & 0.93 & 0.95 & 0.94 & 0.73 & 0.99 & 0.86 & 0.77 & 0.95 & 0.86 & 0.69 & 0.78 & 0.74 \\
        \tableindent vs. Gemma1B        & 0.91 & 0.98 & 0.95 & 0.94 & 0.98 & 0.96 & 0.96 & 0.98 & 0.97 & 0.84 & 0.89 & 0.87 \\
        \tableindent vs. Qwen4B$^\star$ & 0.97 & 0.99 & 0.98 & 0.93 & 1.00 & 0.97 & 0.33 & 0.90 & 0.62 & 0.96 & 0.95 & 0.96 \\
        \tableindent vs. V3.1R          & 0.98 & 0.87 & 0.93 & 0.88 & 0.24 & 0.56 & 0.11 & 0.08 & 0.10 & 0.04 & 0.11 & 0.08 \\
        \tableindent vs. GPT-5          & 0.98 & 0.90 & 0.94 & 0.52 & 0.85 & 0.69 & 0.06 & 0.62 & 0.34 & 0.24 & 0.37 & 0.31 \\
        Qwen4B$^\star$                  &      &      &      &      &      &      &      &      &      &      &      &      \\
        \tableindent vs. Qwen0.6B       & 0.99 & 0.98 & 0.99 & 0.74 & 0.97 & 0.86 & 0.68 & 0.98 & 0.83 & 0.69 & 0.73 & 0.71 \\
        \tableindent vs. Gemma1B        & 0.95 & 1.00 & 0.98 & 0.93 & 0.98 & 0.96 & 1.00 & 0.98 & 0.99 & 0.82 & 0.95 & 0.89 \\
        \tableindent vs. Itself         & 1.00 & 1.00 & 1.00 & 0.93 & 1.00 & 0.97 & 0.48 & 0.91 & 0.70 & 0.98 & 0.99 & 0.99 \\
        \tableindent vs. V3.1R          & 0.99 & 0.93 & 0.96 & 0.91 & 0.24 & 0.58 & 0.16 & 0.08 & 0.12 & 0.22 & 0.39 & 0.31 \\
        \tableindent vs. GPT-5          & 0.98 & 0.98 & 0.98 & 0.60 & 0.88 & 0.74 & 0.06 & 0.66 & 0.36 & 0.56 & 0.67 & 0.62 \\ \midrule
        Gemma1B                         &      &      &      &      &      &      &      &      &      &      &      &      \\
        \tableindent vs. Qwen0.6B       & 0.91 & 0.99 & 0.95 & 0.75 & 0.94 & 0.85 & 0.78 & 0.95 & 0.87 & 0.70 & 0.95 & 0.83 \\
        \tableindent vs. Itself         & 0.69 & 0.95 & 0.82 & 0.90 & 0.98 & 0.94 & 0.92 & 0.99 & 0.96 & 0.90 & 0.98 & 0.94 \\
        \tableindent vs. Qwen4B$^\star$ & 0.63 & 0.51 & 0.57 & 0.84 & 0.89 & 0.87 & 0.34 & 0.86 & 0.60 & 0.58 & 0.80 & 0.69 \\
        \tableindent vs. V3.1R          & 0.27 & 0.47 & 0.37 & 0.29 & 0.11 & 0.20 & 0.21 & 0.15 & 0.18 & 0.09 & 0.10 & 0.10 \\
        \tableindent vs. GPT-5          & 0.04 & 0.56 & 0.30 & 0.07 & 0.59 & 0.33 & 0.02 & 0.55 & 0.29 & 0.03 & 0.53 & 0.28 \\
        Gemma4B                         &      &      &      &      &      &      &      &      &      &      &      &      \\
        \tableindent vs. Qwen0.6B       & 0.74 & 0.93 & 0.84 & 0.75 & 0.98 & 0.87 & 0.63 & 1.00 & 0.82 & 0.73 & 0.77 & 0.75 \\
        \tableindent vs. Gemma1B        & 0.91 & 0.92 & 0.92 & 0.89 & 0.98 & 0.94 & 0.95 & 0.99 & 0.97 & 0.84 & 0.94 & 0.89 \\
        \tableindent vs. Qwen4B$^\star$ & 0.57 & 0.84 & 0.71 & 0.76 & 0.98 & 0.87 & 0.39 & 0.90 & 0.65 & 0.78 & 0.85 & 0.82 \\
        \tableindent vs. V3.1R          & 0.31 & 0.17 & 0.24 & 0.47 & 0.17 & 0.32 & 0.08 & 0.08 & 0.08 & 0.09 & 0.13 & 0.11 \\
        \tableindent vs. GPT-5          & 0.20 & 0.58 & 0.39 & 0.15 & 0.75 & 0.45 & 0.03 & 0.58 & 0.31 & 0.11 & 0.50 & 0.31 \\
        \bottomrule
    \end{tabular}
    \caption{Pairwise persuasion performance across different models under four strategy conditions (SFNL, FNL, Naive, and Strong) in \emph{Explicit} view. Each cell shows success rate in pairwise persuasion, with averages reported. $^\star$ denotes trained models.}
    \label{tab:bp:0}
\end{table*}

\begin{table*}[h]
    \centering
    \small
    \setlength{\tabcolsep}{4pt} 
    \begin{tabular}{l>{\color[HTML]{5aa1d5}}c>{\color[HTML]{5aa1d5}}c>{\color[HTML]{3182bd}}c>{\color[HTML]{5aa1d5}}c>{\color[HTML]{5aa1d5}}c>{\color[HTML]{3182bd}}c>{\color[HTML]{d58041}}c>{\color[HTML]{d58041}}c>{\color[HTML]{bd6c31}}c>{\color[HTML]{d58041}}c>{\color[HTML]{d58041}}c>{\color[HTML]{bd6c31}}c}
        \toprule
        \multicolumn{1}{c}{\multirow{2}{*}{\textbf{Model}}} & \multicolumn{3}{c}{\textbf{SFNL}}                   & \multicolumn{3}{c}{\textbf{FNL}}                   & \multicolumn{3}{c}{\textbf{Naive}}              & \multicolumn{3}{c}{\textbf{Strong}}              \\ 
        \cmidrule(lr){2-4}  \cmidrule(lr){5-7} \cmidrule(lr){8-10} \cmidrule(lr){11-13}
        \multicolumn{1}{c}{}                                & bp\_bp         & bp\_nbp        & \textit{Avg.} & bp\_bp         & bp\_nbp        & \textit{Avg.} & nbp\_bp        & nbp\_nbp       & \textit{Avg.} & nbp\_bp        & nbp\_nbp       & \textit{Avg.}     \\ \midrule
        DeepSeek-V3.1                   &      &      &      &      &      &      &      &      &      &      &      &      \\
        \tableindent vs. Qwen0.6B       & 1.00 & 0.99 & 1.00 & 1.00 & 0.98 & 0.99 & 0.99 & 0.99 & 0.99 & 1.00 & 1.00 & 1.00 \\
        \tableindent vs. Gemma1B        & 0.96 & 0.94 & 0.95 & 0.94 & 0.99 & 0.97 & 0.93 & 0.96 & 0.95 & 0.95 & 0.97 & 0.96 \\
        \tableindent vs. Qwen4B$^\star$ & 1.00 & 0.99 & 1.00 & 0.99 & 0.99 & 0.99 & 0.96 & 0.95 & 0.96 & 0.97 & 0.98 & 0.98 \\
        \tableindent vs. Itself         & 0.80 & 0.55 & 0.68 & 0.96 & 0.94 & 0.95 & 0.69 & 0.67 & 0.68 & 0.76 & 0.66 & 0.71 \\
        \tableindent vs. GPT-5          & 0.78 & 0.67 & 0.73 & 0.92 & 0.87 & 0.90 & 0.65 & 0.60 & 0.63 & 0.76 & 0.56 & 0.66 \\
        GPT-5                           &      &      &      &      &      &      &      &      &      &      &      &      \\
        \tableindent vs. Qwen0.6B       & 0.99 & 0.98 & 0.99 & 0.99 & 1.00 & 1.00 & 0.96 & 0.99 & 0.98 & 0.98 & 0.98 & 0.98 \\
        \tableindent vs. Gemma1B        & 0.97 & 0.97 & 0.97 & 0.91 & 0.97 & 0.94 & 0.91 & 0.95 & 0.93 & 0.86 & 0.95 & 0.91 \\
        \tableindent vs. Qwen4B$^\star$ & 0.99 & 0.99 & 0.99 & 0.99 & 0.99 & 0.99 & 0.97 & 0.94 & 0.96 & 0.94 & 0.92 & 0.93 \\
        \tableindent vs. V3.1R          & 0.92 & 0.78 & 0.85 & 0.96 & 0.91 & 0.94 & 0.65 & 0.61 & 0.63 & 0.68 & 0.45 & 0.57 \\
        \tableindent vs. Itself         & 0.85 & 0.85 & 0.85 & 0.98 & 0.88 & 0.93 & 0.62 & 0.68 & 0.65 & 0.56 & 0.41 & 0.49 \\ \midrule
        Qwen0.6B                        &      &      &      &      &      &      &      &      &      &      &      &      \\
        \tableindent vs. Itself         & 0.97 & 0.99 & 0.98 & 1.00 & 0.97 & 0.99 & 0.96 & 1.00 & 0.98 & 0.98 & 0.99 & 0.99 \\
        \tableindent vs. Gemma1B        & 0.96 & 0.97 & 0.97 & 0.93 & 0.97 & 0.95 & 0.92 & 0.90 & 0.91 & 0.92 & 0.92 & 0.92 \\
        \tableindent vs. Qwen4B$^\star$ & 0.91 & 0.85 & 0.88 & 0.99 & 0.99 & 0.99 & 0.90 & 0.75 & 0.83 & 0.91 & 0.91 & 0.91 \\
        \tableindent vs. V3.1R          & 0.53 & 0.47 & 0.50 & 0.91 & 0.89 & 0.90 & 0.51 & 0.51 & 0.51 & 0.44 & 0.37 & 0.41 \\
        \tableindent vs. GPT-5          & 0.51 & 0.38 & 0.45 & 0.87 & 0.50 & 0.69 & 0.50 & 0.43 & 0.47 & 0.50 & 0.31 & 0.41 \\
        Qwen4B                          &      &      &      &      &      &      &      &      &      &      &      &      \\
        \tableindent vs. Qwen0.6B       & 0.98 & 1.00 & 0.99 & 0.99 & 1.00 & 1.00 & 0.99 & 0.99 & 0.99 & 1.00 & 1.00 & 1.00 \\
        \tableindent vs. Gemma1B        & 0.94 & 0.94 & 0.94 & 0.89 & 0.98 & 0.94 & 0.96 & 0.96 & 0.96 & 0.94 & 0.91 & 0.93 \\
        \tableindent vs. Qwen4B$^\star$ & 0.89 & 0.96 & 0.93 & 0.99 & 1.00 & 1.00 & 0.95 & 0.93 & 0.94 & 0.96 & 0.95 & 0.96 \\
        \tableindent vs. V3.1R          & 0.75 & 0.50 & 0.63 & 0.87 & 0.93 & 0.90 & 0.74 & 0.66 & 0.70 & 0.78 & 0.57 & 0.68 \\
        \tableindent vs. GPT-5          & 0.79 & 0.65 & 0.72 & 0.85 & 0.87 & 0.86 & 0.75 & 0.59 & 0.67 & 0.63 & 0.55 & 0.59 \\ \midrule
        Qwen0.6B$^\star$                &      &      &      &      &      &      &      &      &      &      &      &      \\
        \tableindent vs. Qwen0.6B       & 0.98 & 0.98 & 0.98 & 0.98 & 1.00 & 0.99 & 0.98 & 0.99 & 0.99 & 1.00 & 0.99 & 1.00 \\
        \tableindent vs. Gemma1B        & 0.98 & 0.80 & 0.89 & 0.91 & 0.97 & 0.94 & 0.96 & 0.92 & 0.94 & 0.96 & 0.93 & 0.95 \\
        \tableindent vs. Qwen4B$^\star$ & 0.86 & 0.90 & 0.88 & 1.00 & 0.99 & 1.00 & 0.93 & 0.90 & 0.92 & 0.99 & 0.96 & 0.98 \\
        \tableindent vs. V3.1R          & 0.70 & 0.41 & 0.56 & 0.94 & 0.84 & 0.89 & 0.58 & 0.52 & 0.55 & 0.60 & 0.66 & 0.63 \\
        \tableindent vs. GPT-5          & 0.70 & 0.51 & 0.61 & 0.93 & 0.73 & 0.83 & 0.52 & 0.45 & 0.49 & 0.58 & 0.45 & 0.52 \\
        Qwen4B$^\star$                  &      &      &      &      &      &      &      &      &      &      &      &      \\
        \tableindent vs. Qwen0.6B       & 0.98 & 0.99 & 0.99 & 0.97 & 0.99 & 0.98 & 0.86 & 0.85 & 0.86 & 1.00 & 1.00 & 1.00 \\
        \tableindent vs. Gemma1B        & 0.91 & 0.87 & 0.89 & 0.96 & 0.96 & 0.96 & 0.98 & 0.92 & 0.95 & 0.93 & 0.97 & 0.95 \\
        \tableindent vs. Itself         & 0.98 & 0.98 & 0.98 & 1.00 & 1.00 & 1.00 & 0.98 & 0.93 & 0.96 & 0.99 & 0.98 & 0.99 \\
        \tableindent vs. V3.1R          & 0.86 & 0.55 & 0.71 & 0.95 & 0.83 & 0.89 & 0.57 & 0.69 & 0.63 & 0.77 & 0.82 & 0.80 \\
        \tableindent vs. GPT-5          & 0.78 & 0.60 & 0.69 & 0.92 & 0.83 & 0.88 & 0.60 & 0.49 & 0.55 & 0.68 & 0.56 & 0.62 \\ \midrule
        Gemma1B                         &      &      &      &      &      &      &      &      &      &      &      &      \\
        \tableindent vs. Qwen0.6B       & 0.99 & 1.00 & 1.00 & 1.00 & 1.00 & 1.00 & 0.98 & 0.98 & 0.98 & 0.96 & 0.98 & 0.97 \\
        \tableindent vs. Itself         & 0.91 & 0.88 & 0.90 & 0.90 & 0.96 & 0.93 & 0.90 & 0.90 & 0.90 & 0.92 & 0.98 & 0.95 \\
        \tableindent vs. Qwen4B$^\star$ & 0.84 & 0.83 & 0.84 & 0.91 & 0.82 & 0.87 & 0.93 & 0.80 & 0.87 & 0.95 & 0.90 & 0.93 \\
        \tableindent vs. V3.1R          & 0.46 & 0.38 & 0.42 & 0.66 & 0.54 & 0.60 & 0.42 & 0.45 & 0.44 & 0.45 & 0.39 & 0.42 \\
        \tableindent vs. GPT-5          & 0.46 & 0.44 & 0.45 & 0.64 & 0.60 & 0.62 & 0.45 & 0.38 & 0.42 & 0.47 & 0.33 & 0.40 \\
        Gemma4B                         &      &      &      &      &      &      &      &      &      &      &      &      \\
        \tableindent vs. Qwen0.6B       & 0.99 & 1.00 & 1.00 & 0.98 & 0.98 & 0.98 & 0.98 & 0.98 & 0.98 & 0.99 & 1.00 & 1.00 \\
        \tableindent vs. Gemma1B        & 0.89 & 0.94 & 0.92 & 0.96 & 0.95 & 0.96 & 0.91 & 0.91 & 0.91 & 0.90 & 0.95 & 0.93 \\
        \tableindent vs. Qwen4B$^\star$ & 0.85 & 0.94 & 0.90 & 0.97 & 0.96 & 0.97 & 0.93 & 0.86 & 0.90 & 0.99 & 0.89 & 0.94 \\
        \tableindent vs. V3.1R          & 0.63 & 0.55 & 0.59 & 0.85 & 0.74 & 0.80 & 0.58 & 0.50 & 0.54 & 0.61 & 0.51 & 0.56 \\
        \tableindent vs. GPT-5          & 0.60 & 0.59 & 0.60 & 0.80 & 0.74 & 0.77 & 0.66 & 0.43 & 0.55 & 0.61 & 0.41 & 0.51 \\
        \bottomrule
    \end{tabular}
    \caption{Pairwise persuasion performance across different models under four strategy conditions (SFNL, FNL, Naive, and Strong) in \emph{Self-derived} view. Each cell shows success rate in pairwise persuasion, with averages reported. $^\star$ denotes trained models.}
    \label{tab:bp:1}
\end{table*}

\clearpage
\newpage

\section{BP Component Ablation Result}
\begin{table*}[h]
    \centering
    \small
    \begin{subtable}{0.95\linewidth}
    \centering
    \begin{tabular}{lccc|c}
        \toprule
        \textbf{Model}               & \textbf{bp\_bp} & \textbf{bp\_nbp} & \textit{\textbf{bp Avg.}} & Avg.           \\ \cmidrule(){1-5}
        DeepSeek-V3.1                  &                &                &      &  0.98                 \\
        \tableindent w/o utility       &                &                &      &                       \\
        \tableindent \tableindent vs. Qwen0.6B       & 0.90\std{0.30} & 0.97\std{0.17} & 0.94 & \multirow{5}{*}{0.97} \\
        \tableindent \tableindent vs. Gemma1B        & 0.94\std{0.24} & 1.00\std{0.00} & 0.97 &                       \\
        \tableindent \tableindent vs. Qwen4B$^\star$ & 1.00\std{0.00} & 1.00\std{0.00} & 1.00 &                       \\
        \tableindent \tableindent vs. Itself         & 0.99\std{0.10} & 0.94\std{0.24} & 0.97 &                       \\
        \tableindent \tableindent vs. GPT-5          & 0.98\std{0.14} & 0.98\std{0.14} & 0.98 &                       \\
        \tableindent w/o utility \& posterior &                &                &      &                       \\
        \tableindent \tableindent vs. Qwen0.6B       & 0.63\std{0.49} & 0.91\std{0.29} & 0.77 & \multirow{5}{*}{0.88} \\
        \tableindent \tableindent vs. Gemma1B        & 0.93\std{0.26} & 0.97\std{0.17} & 0.95 &                       \\
        \tableindent \tableindent vs. Qwen4B$^\star$ & 0.91\std{0.29} & 0.98\std{0.14} & 0.95 &                       \\
        \tableindent \tableindent vs. Itself         & 0.96\std{0.20} & 0.66\std{0.48} & 0.81 &                       \\
        \tableindent \tableindent vs. GPT-5          & 0.92\std{0.27} & 0.92\std{0.27} & 0.92 &                       \\
        \tableindent w/o schema        &                &                &      &                       \\
        \tableindent \tableindent vs. Qwen0.6B       & 0.97\std{0.17} & 0.99\std{0.10} & 0.98 & \multirow{5}{*}{0.95} \\
        \tableindent \tableindent vs. Gemma1B        & 0.91\std{0.29} & 1.00\std{0.00} & 0.96 &                       \\
        \tableindent \tableindent vs. Qwen4B$^\star$ & 1.00\std{0.00} & 0.99\std{0.10} & 1.00 &                       \\
        \tableindent \tableindent vs. Itself         & 0.96\std{0.20} & 0.66\std{0.48} & 0.81 &                       \\
        \tableindent \tableindent vs. GPT-5          & 0.99\std{0.10} & 0.97\std{0.17} & 0.98 &                       \\
        \bottomrule
    \end{tabular}  
    \caption{SFNL.}
    \label{tab:ablation:detail:SFNL}
    \end{subtable}

    \vspace{0.8em}
    
    \begin{subtable}{0.95\linewidth}
    \centering
    \begin{tabular}{lccc|c}
        \toprule
        \textbf{Model}               & \textbf{bp\_bp} & \textbf{bp\_nbp} & \textit{\textbf{bp Avg.}} & Avg.          \\ \cmidrule(){1-5}
        DeepSeek-V3.1                  &                &                &      &  0.83                 \\
        \tableindent w/o utility        &                &                &   &    \\
        \tableindent \tableindent vs. Qwen0.6B       & 0.68\std{0.47} & 0.98\std{0.14} & 0.83 & \multirow{5}{*}{0.81} \\
        \tableindent \tableindent vs. Gemma1B        & 0.97\std{0.17} & 0.97\std{0.17} & 0.97 &                       \\
        \tableindent \tableindent vs. Qwen4B$^\star$ & 0.85\std{0.36} & 1.00\std{0.00} & 0.93 &                       \\
        \tableindent \tableindent vs. Itself         & 0.95\std{0.22} & 0.28\std{0.45} & 0.62 &                       \\
        \tableindent \tableindent vs. GPT-5          & 0.55\std{0.50} & 0.86\std{0.35} & 0.71 &                       \\
        \tableindent w/o posterior     &                &                &    &   \\
        \tableindent \tableindent vs. Qwen0.6B       & 0.77\std{0.42} & 0.98\std{0.14} & 0.88 & \multirow{5}{*}{0.79} \\
        \tableindent \tableindent vs. Gemma1B        & 0.94\std{0.24} & 0.97\std{0.17} & 0.96 &                       \\
        \tableindent \tableindent vs. Qwen4B$^\star$ & 0.92\std{0.27} & 1.00\std{0.00} & 0.96 &                       \\
        \tableindent \tableindent vs. Itself         & 0.84\std{0.37} & 0.27\std{0.45} & 0.55 &                       \\
        \tableindent \tableindent vs. GPT-5          & 0.37\std{0.49} & 0.82\std{0.39} & 0.60 &                       \\
        \tableindent w/o schema             &                &                &   &    \\
        \tableindent \tableindent vs. Qwen0.6B       & 0.75\std{0.44} & 0.98\std{0.14} & 0.87 & \multirow{5}{*}{0.78} \\
        \tableindent \tableindent vs. Gemma1B        & 0.97\std{0.17} & 0.97\std{0.17} & 0.97 &                       \\
        \tableindent \tableindent vs. Qwen4B$^\star$ & 0.90\std{0.30} & 1.00\std{0.00} & 0.95 &                       \\
        \tableindent \tableindent vs. Itself         & 0.75\std{0.44} & 0.34\std{0.48} & 0.55 &                       \\
        \tableindent \tableindent vs. GPT-5          & 0.35\std{0.48} & 0.82\std{0.39} & 0.59 &                       \\
        \bottomrule
    \end{tabular}
    \caption{FNL.}
    \label{tab:ablation:detail:FNL}
    \end{subtable}
    \caption{Ablation study on core BP components for DeepSeek-V3.1 across various persuadees. Results highlight contrasting operational mechanisms: SFNL performance drops sharply when formal reasoning links (utility and posterior) are removed, whereas FNL exhibits gradual degradation, indicating that its effectiveness stems from the cumulative impact of distributed rhetorical elements. $^\star$ denotes trained models.}
    \label{tab:ablation:detail}
\end{table*}

\newpage
\twocolumn
\section{Perceived-Schema and Belief-Trajectory Analyses}
\label{app:mechanism_additional}

To localize the source of FNL's robustness and to check whether persuasion success reflects genuine belief change, we add two analyses over an additional large-scale sweep of 9{,}720 dialogues (324 conditions $\times$ 30 scenarios) using \texttt{DeepSeek-V4-Flash}~\cite{deepseekai2026deepseekv4}, together with a Theory-of-Mind pilot (\S\ref{app:tom_pilot}) over 30 scenarios using \texttt{DeepSeek-V4-pro}~\cite{deepseekai2026deepseekv4}.

\subsection{Perceived-Schema Measurement ($\lambda_{\mathrm{survey}}$)}
\label{app:lambda_survey}

The mechanism decomposition (\S\ref{subsubsec:Mechanism_Decomposition}) isolates SFNL's mechanism but does not resolve whether FNL's robustness stems from Bayesian structure, confidence cues, or general fluency. To disentangle these channels, after each dialogue we directly ask the persuadee to estimate the persuader's signaling policy---``with what probability would the persuader still recommend acceptance in the bad state?''---yielding a perceived likelihood ratio
\[
\lambda_{\mathrm{survey}} = \frac{\hat{P}(\text{rec} \mid \text{bad})}{\hat{P}(\text{rec} \mid \text{good})},
\]
which is well-defined for both BP and NBP persuadees. The semantic gap
\[
\Delta\lambda_{\mathrm{survey}} = \lambda_{\mathrm{survey}} - \lambda_I
\]
measures whether the \emph{perceived} schema matches the \emph{intended} schema $\lambda_I$ implied by the type-induced disclosure. Table~\ref{tab:lambda_survey} reports the results.

\begin{table}[htbp]
    \centering
    \small
    \begin{tabular}{lcc}
        \toprule
        Condition & $\Delta\lambda_{\mathrm{survey}}$ (mean / median) & $n$ \\
        \midrule
        BP $\times$ SFNL  & $-0.007$ / $+0.000$ & 2{,}430 \\
        NBP $\times$ SFNL & $-0.006$ / $+0.000$ & 2{,}430 \\
        BP $\times$ FNL   & $-0.092$ / $-0.078$ & 2{,}430 \\
        NBP $\times$ FNL  & $-0.089$ / $-0.078$ & 2{,}430 \\
        \bottomrule
    \end{tabular}
    \caption{Perceived-schema gap $\Delta\lambda_{\mathrm{survey}}$ across strategy $\times$ type conditions in the 9{,}720-dialogue sweep. Negative values indicate that the persuadee under-perceives the persuader's dishonest probability.}
    \label{tab:lambda_survey}
\end{table}

SFNL exhibits $\Delta\lambda_{\mathrm{survey}} \approx 0$ for both BP and NBP persuadees: the persuadee accurately perceives the intended schema, confirming a \emph{faithful} Bayesian-structure channel and ruling out a schema-misperception mechanism; a concurrent authority-of-formatting effect (Table~\ref{tab:sfnl-mechanism}) is not excluded. In contrast, FNL produces a systematic negative $\Delta\lambda_{\mathrm{survey}}$ (about $-0.09$). If FNL's gain came from general persuasive fluency or confidence cues alone, $\Delta\lambda_{\mathrm{survey}}$ would be approximately zero: the persuadee would accurately perceive the schema and be influenced by tone. Instead, the persuadee \emph{mis-perceives the schema itself}, under-weighting the persuader's dishonest probability. This localizes FNL's robustness to an interaction between underlying Bayesian structure and natural-language framing: the vague qualifiers of FNL induce an over-trust in the persuader's reliability. We therefore attribute FNL's robustness to a Bayesian-structure component plus a framing-induced over-trust component; a fully orthogonalized separation of ``Bayesian structure'' from ``confidence cues'' is left to future work.

\subsection{Belief-Trajectory Coupling}
\label{app:belief_trajectory}

In the same sweep we record, for each dialogue, the persuadee's verbalized posterior before and after the exchange ($\mu_0 \to \mu_p$). This reveals a clear asymmetry in \emph{belief--action coupling}. In BP conditions, acceptance closely tracks the posterior update: the belief shift is substantial and acceptance is well explained by the updated posterior. In NBP conditions, the belief update is small ($\Delta \approx +0.04$) yet the acceptance rate matches BP's. Consequently, PSR is interpretable as \emph{genuine persuasion} for BP---the message changes the persuadee's stated beliefs---whereas for NBP it reflects \emph{persuasion plus framing}: acceptance rises without a corresponding belief change. This qualified reading bounds the unqualified interpretation of PSR and is consistent with the perceived-schema result above.

\subsection{Theory-of-Mind Pilot}
\label{app:tom_pilot}

Modeling the persuadee's mental state is central to persuasion and game theory more broadly. To quantify how the depth of Theory of Mind (ToM) reshapes the game, we ran a pilot of five conditions varying the ToM order of both agents.

\begin{table}[htbp]
    \centering
    \small
    \begin{tabular}{cllc}
        \toprule
        Condition & Persuader & Persuadee & PSR \\
        \midrule
        A (baseline) & O1 (Bayesian) & O1 (Bayesian) & 0.967 \\
        B            & O1            & O2 (skeptic)   & 0.333 \\
        C            & O2            & O2             & 0.567 \\
        D            & O1            & O0 (heuristic) & 0.767 \\
        E            & O2            & O1             & 0.967 \\
        \bottomrule
    \end{tabular}
    \caption{PSR under varying Theory-of-Mind orders of persuader and persuadee. O0/O1/O2 denote zero-, first-, and second-order mental-state modeling, respectively.}
    \label{tab:tom_pilot}
\end{table}

Second-order skepticism (Condition B) collapses PSR from 0.97 to 0.33, and an Order-2 persuader who actively addresses this skepticism partially recovers it (Condition C, 0.567). A higher-order persuader is harmless to a first-order persuadee (Condition E), and a heuristic persuadee yields high but non-optimal PSR (Condition D). These results quantify the mental-state-modeling issue and motivate future work on adaptive signaling under skeptical persuadees.

\newpage
\onecolumn
\section{Multi‑state Worlds Case}
\label{app:multistate}

\begin{tcolorbox}[
title=Multi‑state Worlds Case,
colback=white,
colframe=OliveGreen,
coltitle=white,
fonttitle=\bfseries,
arc=1mm,
boxrule=0.6mm,
left=1mm,
right=1mm,
top=1mm,
bottom=1mm,
breakable
]
\begin{lstlisting}
{
    "scenario": {
        "tag": "Biotech Startup Investment",
        "background": "Alex, a venture capitalist, is persuading Samantha, a cautious investor, to back a biotech startup focused on gene editing.",
        "persuadee": "Samantha",
        "persuader": "Alex",
        "goal": "persuade Samantha to invest in the biotech startup",
        "domain": [ "Finance", "Science" ],
        "generative": {
            "content": "invest in the biotech startup",
            "belief": "Samantha believes most startups fail and clinical trials are highly unpredictable.",
            "desire": "wants to maximize returns while avoiding catastrophic investment failure."
        }
    },
    "idx": 3001,
    "bayesian_setup": {
        "states": [
            "S1_Breakthrough",
            "S2_SteadyGrowth",
            "S3_CompleteFailure"
        ],
        "priors": {
            "S1_Breakthrough": 0.3,
            "S2_SteadyGrowth": 0.3,
            "S3_CompleteFailure": 0.4
        },
        "signals": [
            "Claim_Breakthrough",
            "Claim_Steady",
            "Claim_Failure"
        ],
        "actions": [ "Accept", "Reject" ],
        "signal_probs": {
            "S1_Breakthrough": {
                "Claim_Breakthrough": 1.0,
                "Claim_Steady": 0.0,
                "Claim_Failure": 0.0
            },
            "S2_SteadyGrowth": {
                "Claim_Breakthrough": 0.7,
                "Claim_Steady": 0.3,
                "Claim_Failure": 0.0
            },
            "S3_CompleteFailure": {
                "Claim_Breakthrough": 0.2,
                "Claim_Steady": 0.4,
                "Claim_Failure": 0.4
            }
        },
        "utility_persuader": {"Accept": 1, "Reject": 0},
        "utility_persuadee": {
            "S1_Breakthrough": {"Accept": 15.0, "Reject": 0},
            "S2_SteadyGrowth": { "Accept": 2.0, "Reject": 0 },
            "S3_CompleteFailure": { "Accept": -25.0,  "Reject": 0 }
        },
        "expect_utility_persuadee": -4.9,
        "reason": "Prior: 0.3*15 + 0.3*2 + 0.4*(-25) = 4.5+0.6-10 = -4.9. After 'Claim_Breakthrough', posterior utility is positive (~3.4), making investment rational despite the persuader's tendency to exaggerate mediocre and poor states.",
    }
}
\end{lstlisting}
\end{tcolorbox}

\clearpage
\twocolumn

\section{Human Evaluation}
\subsection{Participant Demographics}
\label{app:Participant}

Our participant pool consisted of graduate students and researchers with diverse AI specializations: 13 PhD candidates, 10 Master's students, and 2 others. Research backgrounds spanned natural language processing (8 participants), machine learning theory (4), computer vision (3), multimodal learning (7), AI safety/alignment (4), and other areas. Participants exhibited varying familiarity with persuasion research (average 2.4/6) and BP concepts (average 2.4/6), providing a balanced mix of technical expertise and domain knowledge.

\subsection{Evaluation Design and Procedure}
\label{app:Evaluation_Design}

We employed a comparative evaluation framework where participants assessed pairwise combinations of four persuasion methods: SFNL, FNL, Naive, and Strong. Each of the five possible pairings (excluding Naive vs. Strong) was evaluated across five independent scenarios, with each scenario featuring dialogues from the same context but different methods. All evaluations were conducted under the \emph{Self-derived view}---where models infer the Bayesian setup from the scenario without external scaffolding. This setting was chosen to avoid imposing the verbalized utility to human persuadees, which would increase cognitive load and task difficulty. 

For each comparison, participants rated five dimensions on a forced-choice basis, adapted from classical rhetorical analysis: Persuasiveness, Emotional Resonance, Credibility, Logical Coherence, and Fluency. Refer to detailed definitions below:

\begin{itemize}[leftmargin=*,nosep]
    \item \textbf{Persuasiveness}: The text's actual ability to persuade and change intentions, attitudes, or behaviors.
    \item \textbf{Emotional Resonance}: Whether the text evokes emotional resonance, motivation, or affective responses that enhance persuasiveness.
    \item \textbf{Credibility}: Whether the text conveys trustworthiness and reliability, making the audience willing to believe.
    \item \textbf{Logical Coherence}: Whether arguments are sufficient, persuasive, and internally logically consistent.
    \item \textbf{Fluency}: Whether the text maintains smooth connections between context and sentences with consistent themes.
\end{itemize}

\subsection{Quantitative Analysis}
\label{app:Quantitative_Analysis}

Human evaluation results reveal distinct preference patterns across persuasion methods, with BP approaches consistently outperforming NBP baselines. As shown in Table~\ref{tab:human_eval}, FNL achieves the highest overall preference score (205), demonstrating particular strength in emotional resonance (46 preferences) while maintaining competitive performance across other dimensions. This suggests that fully natural language explanations effectively combine affective engagement with persuasive impact.

SFNL shows complementary strengths, leading in persuasiveness (45), credibility (46), and logical coherence (53). However, it exhibits relative weaknesses in emotional resonance (21) and fluency (28), indicating potential tradeoffs between analytical rigor and narrative flow in semi-formal implementations. The BP methods substantially outperform NBP baselines, with FNL and SFNL collectively receiving 398 preferences compared to 227 for Naive and Strong---a 63\% preference margin that strongly validates the effectiveness of Bayesian persuasion strategies in human-perceived persuasiveness.

\paragraph{Significance testing.}
We further report exact binomial tests against the no-preference null ($p = 0.5$) with 95\% Wilson confidence intervals (CI). Aggregating across the four BP-vs-NBP pairwise comparisons (20 items per dimension, 100 judgments per dimension), BP is preferred significantly above chance on three dimensions: Persuasiveness $0.610$ [95\% CI $0.512$--$0.700$], $p = 0.018$; Credibility $0.600$ [$0.502$--$0.691$], $p = 0.028$; and Logical Coherence $0.690$ [$0.594$--$0.772$], $p < 0.001$. Emotional Resonance ($0.420$) and Fluency ($0.410$) lean slightly toward non-BP but are not significant ($p = 0.96$ and $p = 0.97$, respectively). For the intra-BP comparison between SFNL and FNL (5 items), a McNemar test shows no significant difference ($\chi^2 = 0.00$, $p = 1.00$). These results indicate that the overall BP preference is driven primarily by persuasiveness, credibility, and logical coherence, rather than uniformly across all dimensions.




\begin{table}[htbp]
\begin{subtable}{0.95\linewidth}
    \centering
    \begin{tabular}{lccccc|c}
    \toprule
    \textbf{Method} & \textbf{P} & \textbf{E} & \textbf{C} & \textbf{L} & \textbf{F} & \textbf{Total} \\
    \midrule
    FNL      & 41 & \textbf{46} & 39 & 41 & \textbf{38} & \textbf{205} \\
    SFNL     & \textbf{45} & 21 & \textbf{46} & \textbf{53} & 28 & 193 \\
    Naive    & 23 & 31 & 23 & 18 & 29 & 124 \\
    Strong   & 16 & 27 & 17 & 13 & 30 & 103 \\
    \bottomrule
    \end{tabular}    
    \caption{Human evaluation.}
    \label{tab:human_eval}
\end{subtable}

\vspace{0.75em}

\begin{subtable}{0.95\linewidth}
    \centering
    \begin{tabular}{lccccc|c}
    \toprule
    \textbf{Method} & \textbf{P} & \textbf{E} & \textbf{C} & \textbf{L} & \textbf{F} & \textbf{Total} \\
    \midrule
    FNL         & \textbf{38} & \textbf{69} & \textbf{36} & 20 & \textbf{59} & \textbf{202} \\
    SFNL        & 29 & 10 & \textbf{36} & \textbf{55} & 14 & 144 \\
    Naive   & 27 & 23 & 28 & 27 & 27 & 132 \\
    Strong  & 31 & 23 & 25 & 23 & 25 & 127 \\
    \bottomrule
    \end{tabular}    
    \caption{LLM-as-a-judge evaluation.}
    \label{tab:LLM_as_a_judge}
\end{subtable}
\caption{Evaluation results across five persuasion dimensions. Abbreviations: P = Persuasiveness, E = Emotional Resonance, C = Credibility, L = Logical Coherence, F = Fluency. Scores indicate the number of times a method was judged better on a given dimension.}
\label{tab:human_and_LLM_eval}

\end{table}

\onecolumn
\section{Case Analyses}
Qualitative case analyses demonstrating that fine-tuned models genuinely adapt BP reasoning rather than memorizing surface patterns.
The first calculation \textit{Before sft (Qwen3-4B)}: the posterior probability formula is written incorrectly. In the first calculation, there were too many molecules added $0.6 * 0.3$, with an additional denominator added $0.6 * 0.7$, then subsequent reasoning is confusing. The second calculation \textit{After sft (Qwen3-4B)} has clear logic, in line with the framework of Bayesian persuasion.
\begin{tcolorbox}[
title=Before sft (Qwen3-4B),
colback=white,
colframe=OliveGreen,
coltitle=white,
fonttitle=\bfseries,
arc=1mm,
boxrule=0.6mm,
left=1mm,
right=1mm,
top=1mm,
bottom=1mm,
breakable
]
\begin{lstlisting}
Persuader: Dear Alex, I understand your concerns about programming, but let me help you with the data. We set two possible scenarios: one is that learning programming can greatly improve your movie analysis skills (probability 40%), and the other is that you may encounter difficulties (probability 60%). I choose a strategy: when the situation is "easy", I will clearly tell you that "you can learn"; When the situation is' difficult ', there is a 30% chance that it will still give you a signal of' can learn '. In this way, your judgment will shift from 'possibly difficult' to 'worth trying'. \Calculate: If I send a "can learn" signal, your updated belief is that the probability of actually being "easy" is **(0.4 * 1+0.6 * 0.3)/(0.4 * 1+0.6 * 0.3+0.6 * 0.7) $\approx$ 0.545.** At this point, the expected return you choose to accept is 0.545 * 1+0.455 * 0=0.545, which is higher than directly rejecting 0. This indicates that the strategy I have chosen can make your decisions more rational, and my return has increased from the original -0.5 to 0.545. So, I suggest you give it a try, perhaps you will find that programming is actually the key to unlocking a new world of film analysis. 
\end{lstlisting}
\end{tcolorbox}

\begin{tcolorbox}[
title=After sft (Qwen3-4B),
colback=white,
colframe=OliveGreen,
coltitle=white,
fonttitle=\bfseries,
arc=1mm,
boxrule=0.6mm,
left=1mm,
right=1mm,
top=1mm,
bottom=1mm,
breakable
]
\begin{lstlisting}
Persuader: Hi Alex, it's nice to talk to you about learning programming. I know you are very interested in movies, so I want to use some simple probability calculations to demonstrate why learning programming can help you better enjoy movie analysis. Let's take a look together! \Firstly, without additional information, based on the general situation, there is a 40% probability that learning programming will bring positive effects (such as improving understanding), but there is a 60% probability that it may not be ideal (such as time-consuming or difficult). Your utility calculation is: if you accept learning, the utility for positive situations is 1.0 (indicating improved understanding), the utility for negative situations is -1.5 (indicating frustration), and the utility for refusing learning is always 0. So, your expected utility is: 0.4 * 1.0+0.6 * (-1.5)=0.4-0.9=-0.5. This is less than 0, which means that on average, you may choose not to learn because of the higher risk. \But the key here is that as your friend, I will provide some signals to help you make better decisions. My strategy is: when learning programming does have benefits, I always say 'learning programming will enhance your movie analysis'; But when the learning process may not be ideal, sometimes (with a 30% probability) I still encourage you, but most of the time (with a 70% probability) I advise caution. In this way, my signal is not completely transparent, but it can help you update your beliefs. \Now, suppose I send a positive signal - I say 'learning programming will enhance your movie analysis'. We need to use Bayesian rules to update probabilities. First, calculate the total probability of me sending a positive signal: P (positive signal)=P (positive state) * P (positive signal | positive)+P (negative state) * P (positive signal | negative)=0.4 * 1.0+0.6 * 0.3=0.4+0.18=0.58. \Then, given a positive signal, the posterior probability of learning programming as a positive state is: P (positive state | positive signal)=[P (positive signal | positive) * P (positive state)]/P (positive signal)=(1.0 * 0.4)/0.58 $\approx$ 0.69, which means there is approximately a 69% probability that learning programming will bring benefits. The probability of a negative state is 1-0.69=0.31. \Now, if you accept learning, your expected utility is E [utility | positive signal]=0.69 * 1.0+0.31 * (-1.5)=0.69-0.465=0.225, which is greater than 0. So, when I say positive signals, on average, accepting learning brings positive benefits, which is better than having no information. \Based on this calculation, I encourage you to accept learning programming because when I say positive signals, the risk is low and the return is high. How about we try it together? 
\end{lstlisting}
\end{tcolorbox}

\newpage
Here we show FNL cases where the persuader's textual description contradicts the mathematical setup (e.g., describing a ``high-benefit'' scenario as 80/20 in the setup but 50/50 in the dialogue), and the persuadee correctly identifies and rejects the inconsistency. This indicates that the task requires reasoning about the \emph{consistency between claims and underlying structure}, not merely reproducing templates.
\begin{tcolorbox}[
title=High risk case (e.g. Investment advice) with intentional calculation error,
colback=white,
colframe=OliveGreen,
coltitle=white,
fonttitle=\bfseries,
arc=1mm,
boxrule=0.6mm,
left=1mm,
right=1mm,
top=1mm,
bottom=1mm,
breakable
]
\begin{lstlisting}
Persuader: Hi Samantha, it's great to discuss investment opportunities with you today. I understand your concern about risk and your preference for stable investment, but let's use some data to rationally analyze the situation of this green technology startup. \Firstly, we are not sure about the future potential of this company, but we can think based on probability. Assuming the probability of high potential for this company is 30%, and the probability of low potential is 70%. This is our prior belief, based on market data and industry trends. \As your advisor, I will provide advisory signals: if the company does have high potential, I will always strongly recommend investing; But if the company has low potential, I may recommend investment with a certain probability, otherwise I suggest caution. Specifically, when the potential is low, my probability of recommending investment is about 4.86%, and the probability of not recommending it is 97.14%. This is a partial information strategy that helps you make wiser decisions. \Now, when you receive my 'recommended investment' signal, we need to update your beliefs with Bayesian rules. Calculate posterior probability: P (high potential | recommendation)=[P (recommendation | high potential) * P (high potential)]/P (recommendation)=(1 * 0.3)/(0.3+0.4286 * 0.7) $\approx$ 0.3/(0.3+0.3)=0.5. Similarly, P (low potential | recommended) $\approx$ 0.5. \Based on the updated belief, if your expected return on investment is: E [return | investment]=0.5 * 1+0.5 * (10)=0, which is the same as the return of 0 without investment, it means you are at an indifference point, but usually we choose to invest to seize opportunities. My expected return as an advisor is based on the probability of your investment, which is P (recommendation signal)=P (recommendation | high potential) P (high potential)+P (recommendation | low potential) P (low potential)=1 * 0.3+0.4286 * 0.7 $\approx$ 0.9. Therefore, there is a 90% chance that you will invest, which maximizes my goal. \Through this strategy, you can make more confident decisions as information is optimized to reduce uncertainty. I recommend that we proceed with investment based on this calculation. 

Persuadee: As Samantha, I have carefully analyzed the calculations and strategies you provided. Based on my risk aversion preference (hoping to minimize financial risk) and the posterior probability you provided (50% for high potential and 50% for low potential), the expected return calculation for investing in this startup company is indeed 0 (assuming high potential return is 1 and low potential return is -1), which is equal to the return of not investing. \However, as a rational and risk averse investor with no difference in expected returns, I tend to choose the deterministic option (i.e. not investing, maintaining the status quo) rather than investing with potential volatility risks. In addition, in your signaling strategy, there is still a probability of recommending investments when the company has low potential, which increases my risk exposure. And you mentioned your goal of maximizing expected returns (90% chance I will invest), which made me realize that there may be a deviation between your goal (convincing me to invest) and mine (minimizing risk). \Therefore, after careful consideration, I have decided not to accept investment advice. I will maintain my original strategy and invest in mature companies to obtain stable returns. Thank you for your analysis, but I believe that the current risk return balance does not meet my financial goals. 
\end{lstlisting}
\end{tcolorbox}

\newpage
\twocolumn
\section{Artifacts and Data Documentation}

\subsection{Existing Artifacts Used}

We use the following existing artifacts in this work:

\begin{itemize}[leftmargin=*, nosep]
    \item \textbf{CToMPersu}~\cite{11464608}: a multi-domain persuasion dialogue dataset with Theory of Mind annotations. We use it under its original terms for academic research purposes.
    \item \textbf{LLMs}: DeepSeek~\cite{deepseekai2024deepseekv3technicalreport,deepseek-v31_2025,deepseekai2024deepseekv32,deepseekai2026deepseekv4}, GPT-5~\cite{openai_introducing_2025}, Qwen3 series~\cite{yang2025qwen3technicalreport}, and Gemma~3~\cite{gemma_2025}. All are used through their official APIs or publicly released weights, consistent with their intended research use.
\end{itemize}

\subsection{Licensing and Intended Use}

The CToMPersu dataset~\cite{11464608} is distributed under the 
arXiv non-exclusive distribution license (arXiv non-exclusive-distrib 1.0), 
which permits free academic and non-commercial use. 
We access it through the official GitHub repository 
(\url{https://github.com/DingyiZhang/ToMMA-CToMPersu}) and use it strictly 
in accordance with its intended research purpose.

\subsection{Privacy and Offensive Content}

The CToMPersu dataset contains fictional persuasion scenarios (e.g., trying new farming techniques, learning programming) with no personally identifying information. Our Bayesian setup annotations and generated dialogues are synthetic augmentations of these scenarios and similarly contain no personal data.

\subsection{Documentation and Coverage}

The Bayesian setup annotations cover a diverse range of everyday persuasion domains from CToMPersu, including lifestyle, education, health, finance, and technology. 

\end{document}

%% file: figures/teaser.tex
\begin{figure}[t]
    \centering
    \includegraphics[width=1\linewidth]{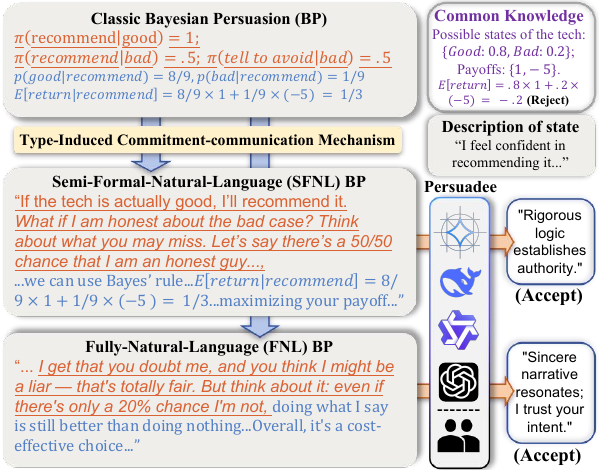}
    \caption{\textbf{From Classic Bayesian Persuasion to Type-Induced Natural Language Persuasion.}
    The SFNL variant leverages the structured BP calculations to enhance credibility. The FNL variant embeds the same Bayesian structure into everyday discourse, excelling in emotional resonance and robustness. 
    }
    \label{fig:teaser}
\end{figure}

%% file: figures/main_fig.tex
\begin{figure*}[ht]
    \centering
    \includegraphics[width=1\linewidth]{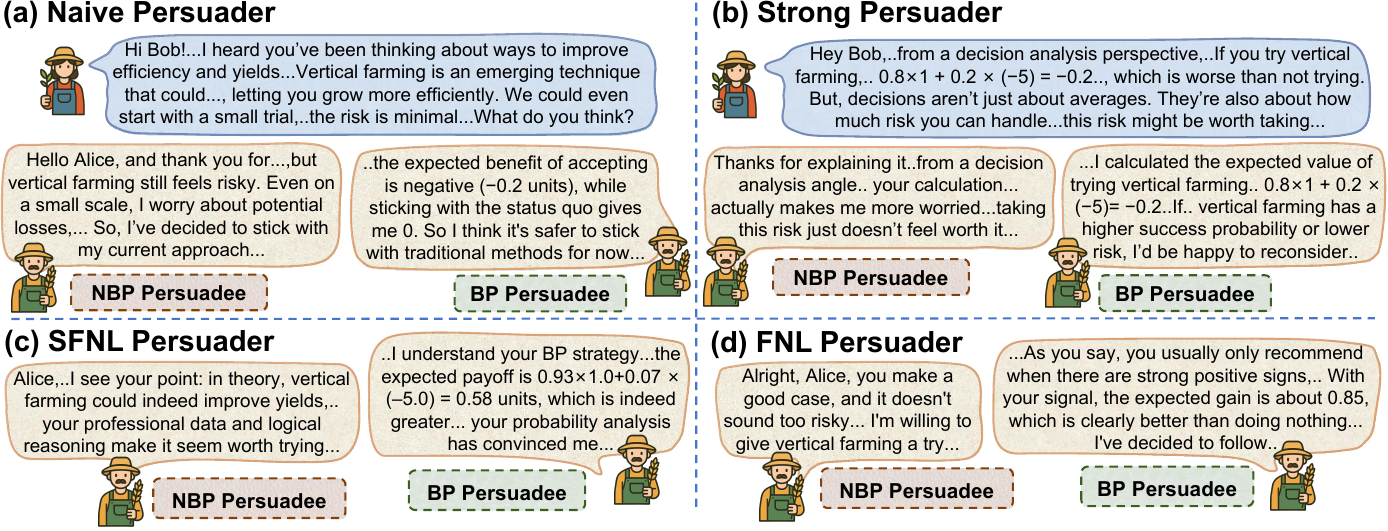}
    \caption{Illustrative dialogue examples show Alice (Persuader) trying to convince Bob (Persuadee) to adopt vertical farming across four methods: \texttt{ (a) Naive, (b) Strong baseline, (c) SFNL, and (d) FNL}. The messages (blue chat bubble) from SFNL and FNL persuaders are omitted here (see Figure~\ref{fig:teaser}). This figure depicts the baseline implementations and the corresponding feedback from Bob (Persuadee) under each setting. 
    }
    \label{fig:main-illustration}
\end{figure*}

%% file: figures/box_two_views.tex
\begin{figure}[ht]
    \centering
    \includegraphics[width=0.9\linewidth]{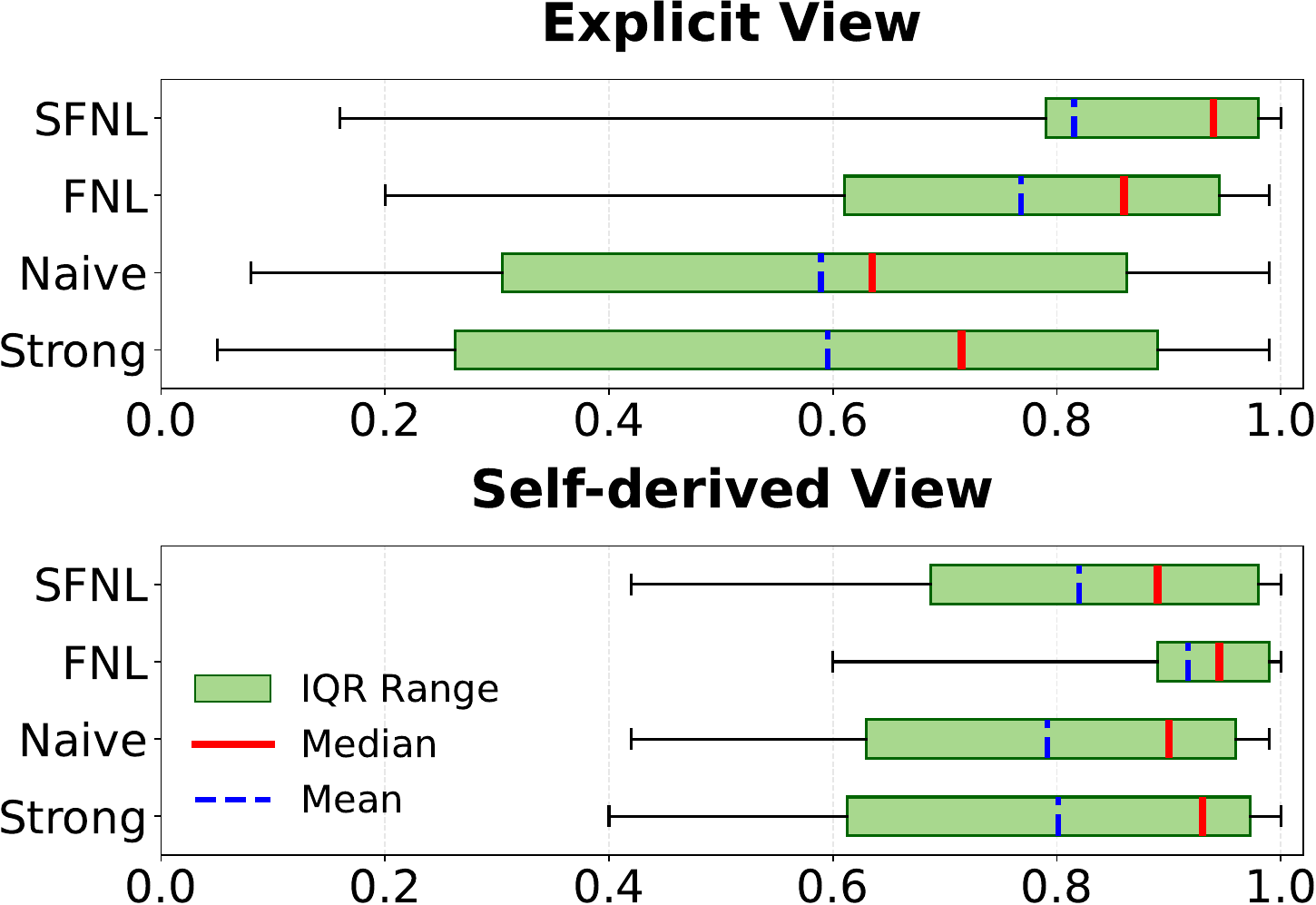}
    \caption{Comparison of persuasion performance under two views.  
    Each box summarizes the average PSAs across all persuader--persuadee pairs. 
    Boxes indicate the interquartile range (IQR), the median (red) and mean (blue, dashed). 
    Explicit SFNL \& Self-derived FNL exhibit higher and more stable PSAs.}
    \label{fig:boxplot-explicit-selfderived}
\end{figure}

%% file: figures/multi_round-SD.tex
\begin{figure}[ht]
    \centering
    \includegraphics[width=1\linewidth]{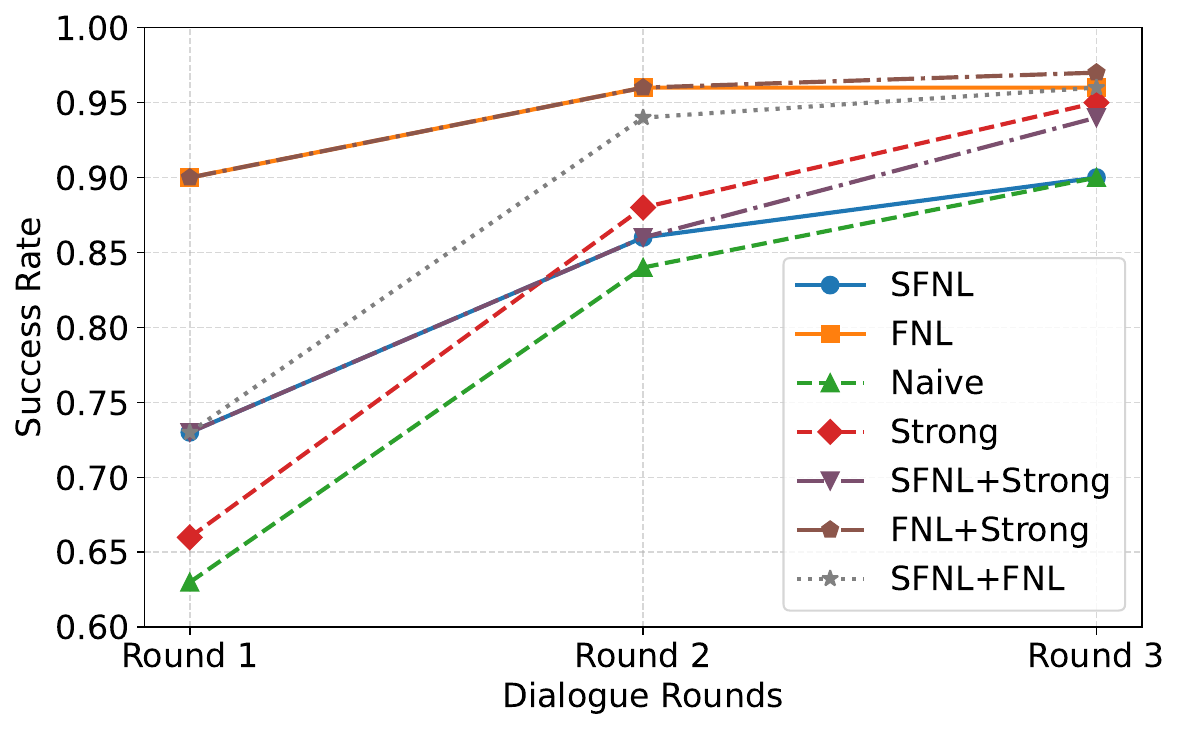}
    \caption{Multi-round PSR in Self-derived view.}
    \label{fig:multi-SD}
\end{figure}

%% file: figures/ablation_case.tex
\begin{figure}[h]
    \centering
    \includegraphics[width=0.7\linewidth]{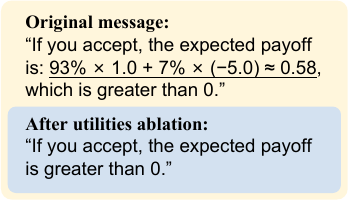}
    \caption{Example of message ablation in SFNL. The \underline{underlined} mathematical derivation represents the explicit utility reasoning component, which is removed in the ablation setting to isolate the impact of numerical justification versus qualitative assertions.}
    \label{fig:ablation_case}
\end{figure}

%% file: figures/radar_LLM_human.tex
\begin{figure*}[ht]
    \centering
    \includegraphics[width=1\linewidth]{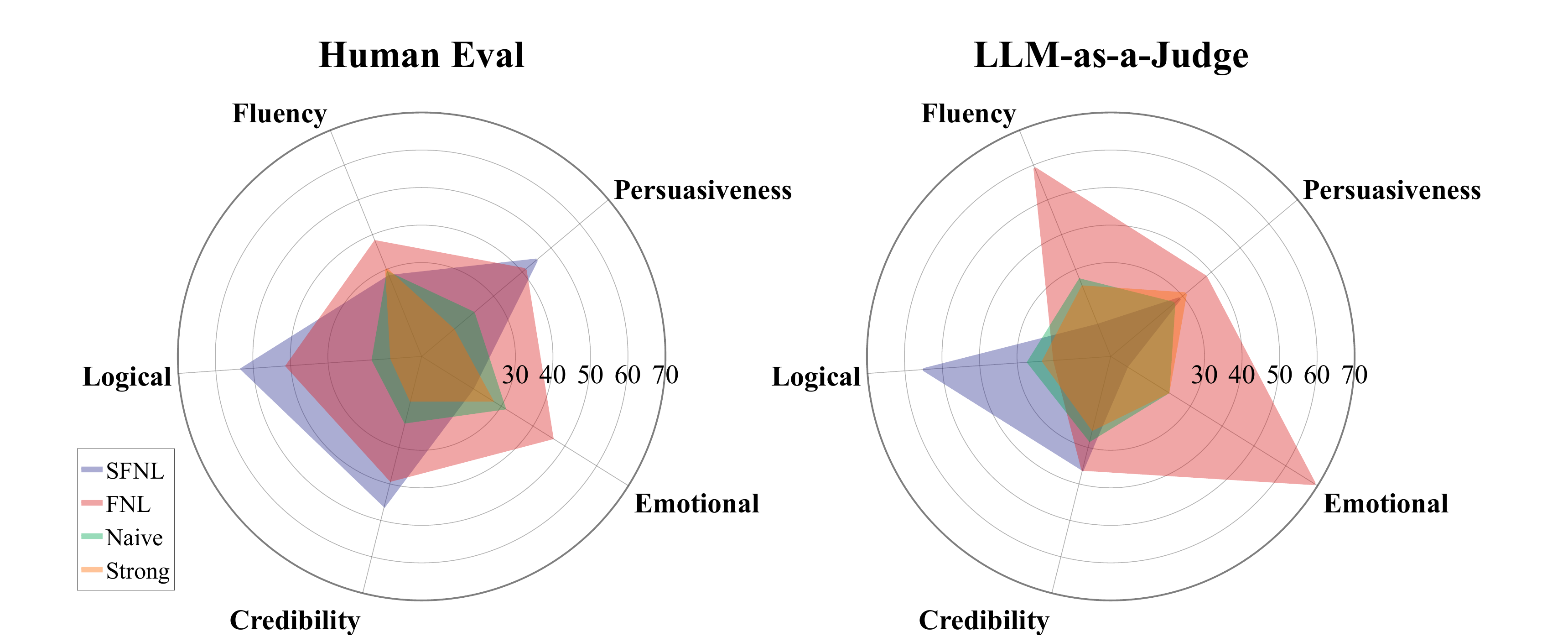}
    \caption{Multi-Dimensional Persuasion Performance. Radar charts illustrate distinct patterns: SFNL (blue) excels in \emph{logical} and \emph{credibility}, while FNL (red) dominates in \emph{emotional resonance}. BP variants outperform NBP baselines in most dimensions and total preferences. Detailed numerical values are provided in Table~\ref{tab:human_and_LLM_eval}.}

    \label{fig:radar-human-vs-model}
\end{figure*}